%% file: main.tex
\documentclass{article} 
\PassOptionsToPackage{numbers,sort&compress}{natbib}

\newif\ificml
\newif\ifneurips
\newif\ificlr
\newif\ifaistats
\icmlfalse
\neuripsfalse
\iclrfalse
\aistatsfalse
\input{math_commands}

\icmltrue \usepackage[accepted]{icml2023}

\usepackage{hyperref}
\usepackage{url}
\usepackage{amsmath,amsfonts,amsthm,amssymb}       
\usepackage{bbm}  
\usepackage[capitalize]{cleveref}  
\usepackage{mathtools}  
\usepackage{algorithm}
\usepackage[algo2e,ruled]{algorithm2e}  
\usepackage{multirow}
\usepackage{booktabs}  
\usepackage{makecell}  
\usepackage{cancel}  
\usepackage{environ}  
\usepackage{color}
\usepackage{colortbl}
\usepackage{xcolor}
\usepackage[inline]{enumitem}
\ifaistats
  \usepackage[authoryear,sort]{natbib}
\fi

\crefname{equation}{Eq.}{Eqs.}
\crefname{figure}{Fig.}{Figs.}
\crefname{section}{Sec.}{Secs.}
\crefname{subsection}{Sec.}{Secs.}
\crefname{appendix}{Appx.}{Appx.}
\crefname{algocf}{Alg.}{Algs.}
\crefname{observation}{Obs.}{Obs.}
\crefname{definition}{Def.}{Defs.}
\crefname{theorem}{Theorem}{Theorems}
\crefname{proposition}{Prop.}{Props.}

\newif\ifcomments
\commentstrue
\ifcomments\newcommand{\comments}[1]{#1}\else\newcommand{\comments}[1]{}\fi
\definecolor{clrgp}{rgb}{.9,0,.9}

\newif\ifrestating
\restatingfalse
\NewEnviron{restatable}[3]{%
  \expandafter\xdef\csname restatethis@#2\endcsname{%
    \unexpanded\expandafter{\BODY}%
  }%
  \begin{#1}[#3]
    \label{#2}
    \BODY
  \end{#1}%
  \newtheorem*{#2}{\Cref{#2} (Restated)}%
}
\newcommand{\restate}[1]{%
  \restatingtrue
  \begin{#1}\csname restatethis@#1\endcsname\end{#1}%
  \restatingfalse
}

\newcommand\titl{GVA: Reconstructing Vivid 3D Gaussian Avatars from Monocular Videos}
\title{\titl}
\author{\authr}

\ificml
  \icmltitlerunning{\titl}
\fi

\begin{document}

\ificml
  \twocolumn[
    \icmltitle{\titl}
    \begin{icmlauthorlist}
      \icmlauthor{Xinqi Liu}{}
      \icmlauthor{Chenming Wu}{}
      \icmlauthor{Jialun Liu}{}
      \icmlauthor{Xing Liu}{}
      \icmlauthor{Jinbo Wu}{} 
      \icmlauthor{Chen Zhao}{}
      \icmlauthor{Haocheng Feng}{}\\
      \icmlauthor{Errui Ding}{}
      \icmlauthor{Jingdong Wang}{}
    \end{icmlauthorlist}
    \vskip 0.3in
  ]
  
  \printAffiliationsAndNotice{\icmlContribution} 
\else
  \ifaistats
    \twocolumn[
      \aistatstitle{\titl}
      \aistatsauthor{\authr}
      \aistatsaddress{ Institution 1 \And  Institution 2 \And Institution 3 }
    ]
  \else
    \maketitle
  \fi
\fi

\begin{abstract}
 In this paper, we present a novel method that facilitates the creation of vivid 3D Gaussian avatars from monocular video inputs (GVA).  Our innovation lies in addressing the intricate challenges of delivering high-fidelity human body reconstructions and aligning 3D Gaussians with human skin surfaces accurately. The key contributions of this paper are twofold. Firstly, we introduce a pose refinement technique to improve hand and foot pose accuracy by aligning normal maps and silhouettes. Precise pose is crucial for correct shape and appearance reconstruction. Secondly, we address the problems of unbalanced aggregation and initialization bias that previously diminished the quality of 3D Gaussian avatars, through a novel surface-guided re-initialization method that ensures accurate alignment of 3D Gaussian points with avatar surfaces. Experimental results demonstrate that our proposed method achieves high-fidelity and vivid 3D Gaussian avatar reconstruction. Extensive experimental analyses validate the performance qualitatively and quantitatively, demonstrating that it achieves state-of-the-art performance in photo-realistic novel view synthesis while offering fine-grained control over the human body and hand pose.  Project page: \href{https://3d-aigc.github.io/GVA/}{https://3d-aigc.github.io/GVA/}
\end{abstract}

\input{1_intro}

\input{2_related_work}

\input{3_method}

\input{4_experiments}

\input{5_conclusion}

{
	\small
  \ificml
    \bibliographystyle{icml2023}
  \else
    \ificlr
      \bibliographystyle{iclr2024_conference}
    \else
      \bibliographystyle{abbrvnat}
    \fi
  \fi
  \bibliography{main}
}

\clearpage
\ificml
  \onecolumn
\else
  \ifaistats
    \onecolumn
  \fi
\fi

\appendix

\end{document}

%% file: math_commands.tex
\usepackage{amsmath,amsfonts,bm}

\def\eqref#1{equation~\ref{#1}}

\def\1{\bm{1}}

\def\vc{{\bm{c}}}

\def\vf{{\bm{f}}}

\def\vq{{\bm{q}}}

\def\vs{{\bm{s}}}

\def\vx{{\bm{x}}}

\DeclareMathAlphabet{\mathsfit}{\encodingdefault}{\sfdefault}{m}{sl}
\SetMathAlphabet{\mathsfit}{bold}{\encodingdefault}{\sfdefault}{bx}{n}



%% file: 1_intro.tex
\section{Introduction}


Reconstructing a drivable and photorealistic avatar from a monocular video or image sequence has garnered considerable attention in academia and industry. This advancement holds tremendous potential for generating substantial commercial value and significantly impacting diverse areas, such as e-commerce marketing, live broadcasting, film production, virtual try-ons, etc.

\begin{figure}[t]
\centering
\includegraphics[width=0.95\linewidth]{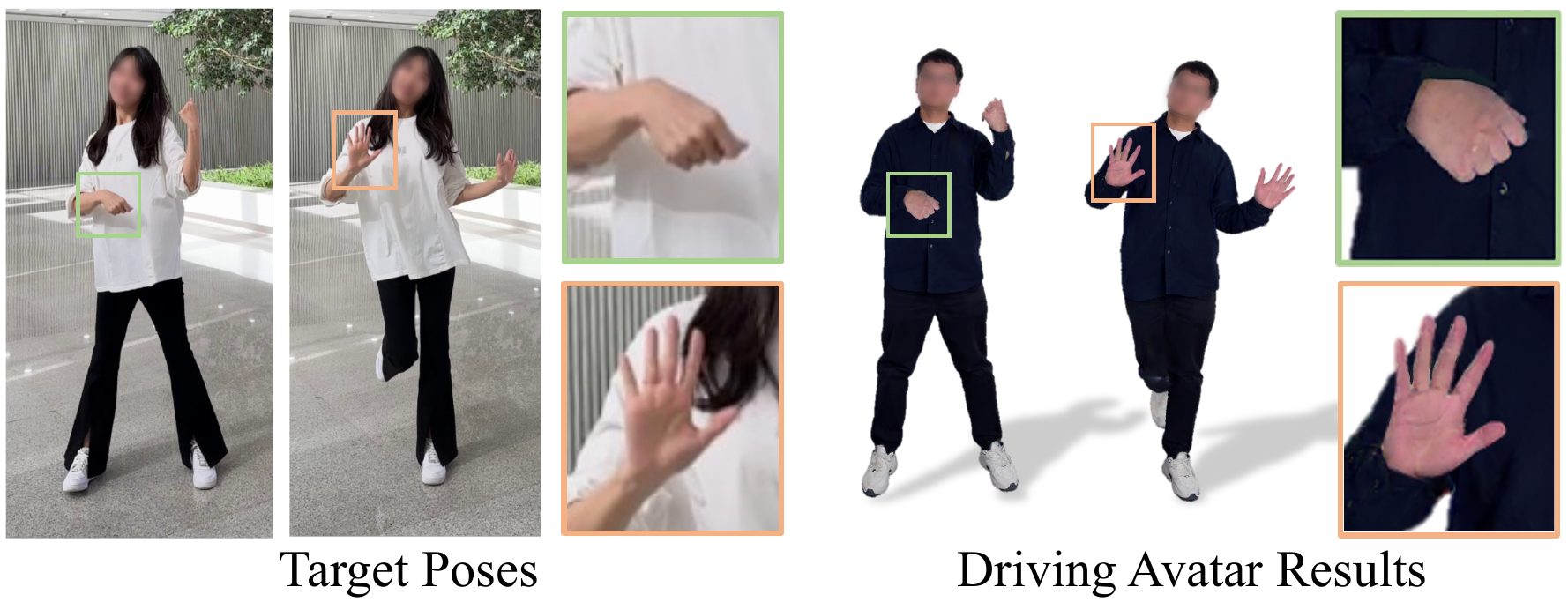}
\caption{Our proposed GVA enables the effective reconstruction of 3D Gaussian avatars from monocular videos. Its capability for flexible pose adjustments via external motions results in realistic avatars.
}
\label{fig1}
\end{figure}

Existing methods for avatar reconstruction heavily rely on RGB-D cameras~\cite{guo2017real,yu2017bodyfusion,yu2018doublefusion}, dome multi-view acquisition equipment~\cite{dou2016fusion4d,guo2019relightables}, or the manual labor of artists to digitally model human subjects, which were then driven using linear blending skinning (LBS) techniques. Nevertheless, these methods encountered challenges related to high costs associated with acquisition and production, as well as struggles in attaining photorealistic rendering results. The advent of Neural Radiation Field (NeRF)\cite{mildenhall2021nerf,barron2021mip} has made it feasible to create cost-effective and photorealistic 3D avatars~\cite{peng2021neural,peng2021animatable,weng2022humannerf,jiang2022selfrecon} leveraging volume rendering techniques. Incorporating the pose-conditioned MLP (Multi-Layer Perceptron) deformation field allows the avatar to be controlled or driven according to specific poses. Despite the favorable qualities exhibited by the neural radiation field, this modeling method encounters challenges such as extensive training durations and limited pose generalization, especially when confronted with significant pose deformations. This is primarily attributed to the inherent implicit representation employed.

\begin{figure*}[t]
\centering
\includegraphics[width=0.95\linewidth]{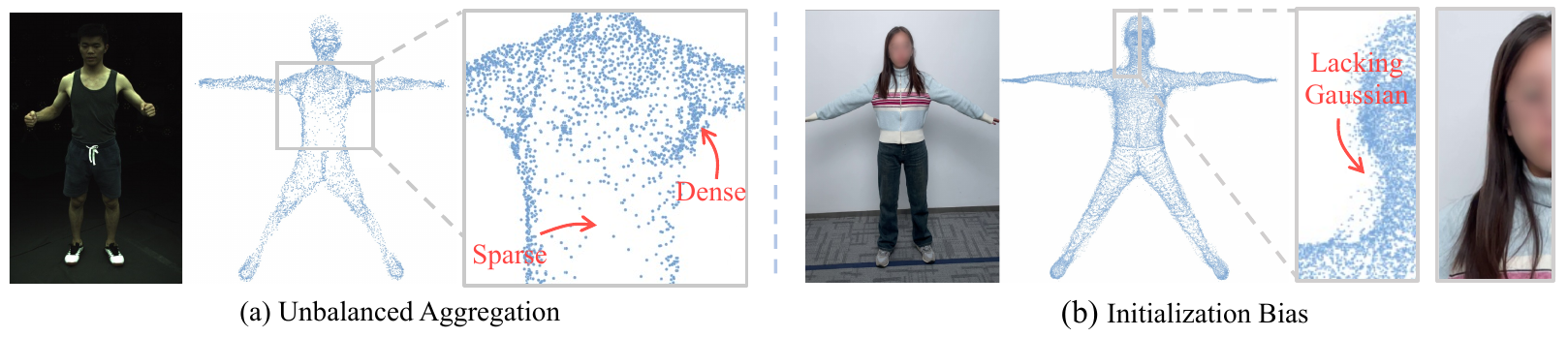}
\vspace{-0.2cm}
\caption{The illustration of the widespread phenomena of unbalanced aggregation and initialization bias within the 3D Gaussian avatar reconstruction algorithms.}
\label{fig2}
\end{figure*}
Recently, 3D Gaussian Splatting (3DGS)~\cite{kerbl20233d} has gained widespread attention due to its explicit representation, remarkable expressiveness, rapid convergence performance, and real-time rendering capabilities. Since its invention, a large amount of work regarding 3D Gaussian avatars has been proposed~\cite{zielonka2023drivable,yuan2023gavatar,qian20233dgs,saito2023relightable,hu2023gauhuman,qian2023gaussianavatars}, achieving unprecedented high-fidelity rendering results combining 3DGS and parametric human models. 

However, those methods encounter two prominent limitations. Firstly, the prevailing avatar model primarily supports body control, lacking the capability to provide expressive functionalities, such as hand-driving. This limitation stems from the inadequate accuracy and stability of whole-body pose predicted by off-the-shelf pose estimation methods such as~\cite{zhang2023pymaf,lin2023one,li2023hybrik}, particularly in the intricate hand and foot regions. 
Secondly, the existing methods exhibit \textbf{unbalanced aggregation} and \textbf{initialization bias} phenomena in building 3D avatars (see Figure~\ref{fig2} for an illustration), leading to potential artifacts in the avatars when driven to novel poses. In particular, dense 3D Gaussian point allocation is observed in high-frequency texture areas, whereas texture-less regions receive a notably sparse point distribution. We refer to this as unbalanced aggregation, as shown in the left part of Figure~\ref{fig2}. Additionally, areas such as shawl hair or accessories that deviate from the initial shape receive less Gaussian point assignment, and we term this as initialization bias, as shown in the right part of Figure~\ref{fig2}. These two properties contribute to an uneven distribution of 3D Gaussian points. They may be beneficial for static scenes but are negative for avatar models. Consequently, even slight deformation in the 3D Gaussian points can significantly impact the rendering outcome, resulting in noticeable artifacts during pose driving.

Our proposed GVA is designed to address the aforementioned challenges. We first introduce a pose refinement by aligning normals and silhouette cues for the first problem. Then, we propose a surface-guided re-initialization mechanism to iteratively redistribute Gaussian points near the surface, alleviating the second problem. As a result, a body and hand-controllable avatar is vividly reconstructed from monocular video, as shown in Figure~\ref{fig1}. The contributions of this paper are summarized as follows.
\begin{itemize}
    \item We propose GVA, a novel method for reconstructing 3D Gaussian avatars directly from monocular video. This method surpasses existing techniques by eliminating the dependency on detailed annotations and showing superior performance in reconstructing avatars within a wide range of settings.
    \item We design a pose refinement method for avatar reconstruction, which significantly improves the accuracy of body and hand alignment, and a surface-guided Gaussian re-initialization mechanism, effectively alleviating unbalanced aggregation and initialization bias issues. 
    \item Extensive experiments have been conducted to validate the effectiveness of our proposed method, proving it can build body- and hand-drivable avatars. 
\end{itemize}



%% file: 2_related_work.tex
\section{Related Work}


\subsection{Human Avatar Reconstruction}
The task of reconstructing avatar models with accurate shapes and realistic appearances has been a long-standing research focus. Early methods typically relied on RGB-D sensors~\cite{izadi2011kinectfusion,newcombe2015dynamicfusion,guo2017real,yu2017bodyfusion,yu2018doublefusion,dou2016fusion4d,dou2017motion2fusion} to capture the shape of the target subject. The reconstructed surface was then manually bound to a predefined skeleton to create the avatar model. However, due to the high cost of scanning and the labor-intensive process of manual skin binding, these methods have not been widely adopted. 
With the development of parametric human models like SMPL~\cite{loper2023smpl} and SMPL-X~\cite{SMPL-X:2019}, low-cost avatar reconstruction becomes possible. This category of approaches allows for the creation of avatars using only RGB images, eliminating the need for expensive scanned data. Many works~\cite{kanazawa2018end,kocabas2020vibe,kolotouros2019convolutional,lin2021end,zhang2023pymaf,lin2023one,li2023hybrik,zhou2021monocular} attempt to estimate the shape and pose parameters of the target subject from images, and then drives the parametric human body model for novel view rendering and novel pose. However, such methods usually solely focus on naked body shapes, lacking user-specific shape details such as clothing. 

Recently, a new pattern of avatar reconstruction appears, which uses a parameterized human body as a priori, and then uses vertex offsets~\cite{ma2020learning,xiang2020monoclothcap}, signed distance field (SDF)~\cite{varol2018bodynet,saito2019pifu,saito2020pifuhd,zheng2021pamir,he2020geo,xiu2022icon,xiu2022econ}, neural radiance field (NeRF)~\cite{kwon2021neural,peng2021neural,peng2021animatable,weng2022humannerf,jiang2022selfrecon,jiang2022neuman} or 3D Gaussian points ~\cite{zielonka2023drivable,yuan2023gavatar,qian20233dgs,saito2023relightable,hu2023gauhuman,qian2023gaussianavatars,jung2023deformable,li2023human101} to enhance the appearance details of user-specific shape features, reconstructing more realistic avatars. Although they significantly enhance the avatar's expressiveness, their reconstruction quality relies heavily on the accuracy of the estimated poses. Existing end-to-end pose estimation methods~\cite{kanazawa2018end,kocabas2020vibe,kolotouros2019convolutional,lin2021end,zhang2023pymaf,lin2023one,li2023hybrik,zhou2021monocular} can only accurately estimate the pure-body pose, while other parts such as hands and foot suffer from obvious misalignment issue. This disadvantage makes avatar reconstruction methods~\cite{kwon2021neural,peng2021neural,peng2021animatable,zielonka2023drivable,yuan2023gavatar,qian20233dgs} only support body-controllable reconstruction. Consequently, these methods face challenges when it comes to directly learning finer-grained controls, such as hand movements.
Instead, our method introduces a pose refinement method for avatar reconstruction, using predicted surface normals and silhouettes as guidance. It significantly reduces the misalignment problem in hand and foot regions, making it possible to easily reconstruct an expressive avatar with a controllable body and hands from monocular videos.

\vspace{-0.4cm}

\subsection{Human Avatar Representation}
The human avatar representation is important for the fidelity and usability of reconstructed avatars. Mesh-based~\cite{loper2023smpl,SMPL-X:2019,ma2020learning,xiang2020monoclothcap,huang2020arch,he2021arch++} and point-cloud-based~\cite{ma2021power} avatar representations are favored over the past few decades due to easy-to-use. However, the discrete nature makes avatars constructed by these methods usually lack high-frequency geometric and texture details. The emergence of NeRF~\cite{mildenhall2021nerf} has motivated many works due to its photorealistic rendering capabilities. NeRF-based representation~\cite{kwon2021neural,peng2021neural,peng2021animatable,weng2022humannerf,jiang2022selfrecon,jiang2022neuman} has achieved unprecedented render quality in novel view. However, this representation usually demands hours of training, and the rendering speed is relatively slow and far from real-time. 

Recently, there has been a surge of interest in the 3D Gaussian splitting (3DGS) representation~\cite{kerbl20233d} due to its ability to achieve a balance between real-time rendering speed and photorealistic rendering quality.
The field of 3D Gaussian-based avatar reconstruction~\cite{zielonka2023drivable,yuan2023gavatar,qian20233dgs,saito2023relightable,hu2023gauhuman,qian2023gaussianavatars,jung2023deformable,li2023human101} has experienced rapid growth and become a bustling area of research within a short period of time. Although these methods effectively exploit the powerful 3D Gaussians for avatar reconstruction, they also inherit harmful properties, such as unbalanced aggregation and initialization bias. This causes the 3D Gaussian-based avatar prone to noticeable artifacts when performing novel pose driving.
Our work also leverages 3D Gaussian representation for avatar reconstruction, and introduces a surface-guided Gaussian re-initialization mechanism to alleviate those issues, improving the avatar's driving ability and expressiveness.

%% file: 3_method.tex
\begin{figure*}[t]
\centering
\includegraphics[width=\linewidth]{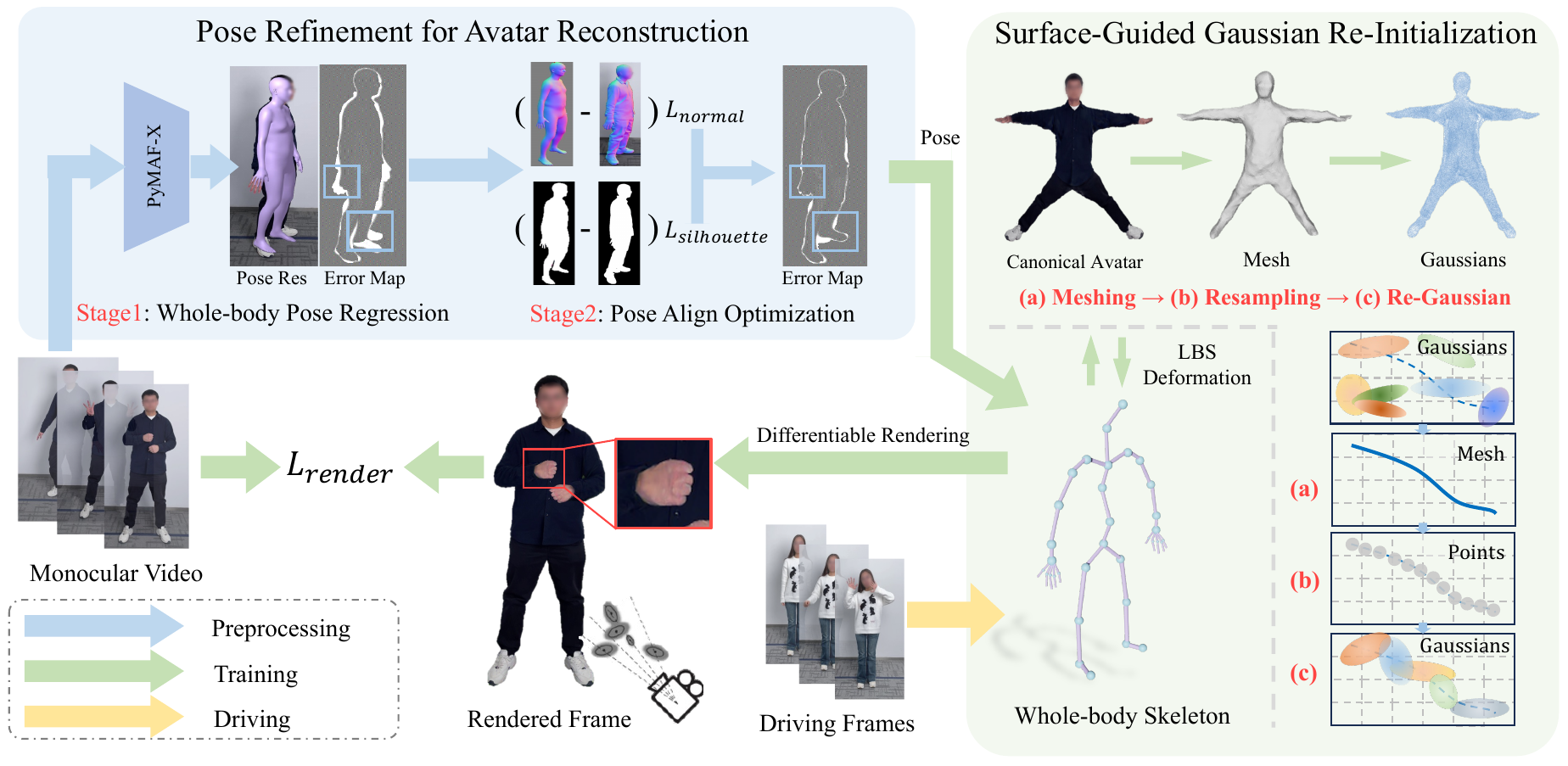}
\vspace{-0.5cm}
\caption{
The framework utilizes a monocular video to obtain refined body and hand poses. The Gaussian avatar model is adjusted based on the whole-body skeleton to match the pose in the image. Consistency with image observations is maintained through differentiable rendering and optimization of Gaussian properties. An surface-guided re-initialization mechanism enhances rendering quality and Gaussian point distribution. The model can adapt to new poses from videos or generated sequences.
}

\label{fig3}
\end{figure*}

\section{Preliminary}

\textbf{3DGS}~\cite{kerbl20233d} employs explicit 3D Gaussian points as its primary rendering entities. A 3D Gaussian point is mathematically defined as a function $G(\vx)$ denoted by
\begin{equation}\label{eq1}
G(\vx) = e^{-\frac{1}{2}(\vx-\bm{\mu})^\intercal \bm{\Sigma}^{-1}(\vx-\bm{\mu})},
\end{equation}
where $\bm \mu$ and $\bm \Sigma$ denote the spatial mean and covariance matrix, respectively. Each Gaussian is also associated with an opacity $\eta$ and a view-dependent color $\vc$ represented by spherical harmonic coefficients $\vf$.
During the rendering process from a specific viewpoint, 3D Gaussians are projected onto the view plane by splatting. The means of these 2D Gaussians are determined using the projection matrix, while the 2D covariance matrices are approximated as
\begin{equation}\label{eq2}
\bm \Sigma' = \bm  J_g  \bm W_g \bm \Sigma \bm W_g^{\top}\bm J_g^{\top},
\end{equation}
where $\bm W_g$ and $\bm J_g$ denote the viewing transformation and the Jacobian of the affine approximation of the perspective projection transformation of Gaussian points.
To obtain the pixel color, alpha-blending is performed on $N$ sequentially layered 2D Gaussians, starting from the front and moving toward the back.\begin{equation}\label{eq3}
C = \sum_{i \in N} T_i \alpha_i \vc_i \quad \text{with} \quad T_i = \prod_{j=1}^{i} (1 - \alpha_j).
\end{equation}
In the splatting process, the opacity factor, denoted by $\alpha$, is computed by multiplying $\eta$ with the contribution of the 2D covariance, calculated from $\bm \Sigma'$ and the pixel coordinate in image space.
The covariance matrix $\bm \Sigma$ is parameterized using a unit quaternion $\vq$ and a 3D scaling vector $\vs$ to ensure a meaningful interpretation during optimization.

\vspace{4pt} \noindent \textbf{Parameterized SMPL-X Model}~\cite{SMPL-X:2019} is an extension of the original SMPL body model~\cite{loper2023smpl} with face and hand, designed to capture more detailed and expressive human deformations.
SMPL-X expands the joint set of SMPL, including those for the face, fingers, and toes. This allows for a more accurate representation of intricate body movements. SMPL-X is defined by a function $ M(\theta, \beta, \psi): \mathbb{R}^{|\theta| \times|\beta| \times|\psi|} \longrightarrow \mathbb{R}^{3 \mathrm{N}}$, parametrized by the pose $\theta \in \mathbb{R}^{3K}$ ($K$ denotes the number of body joints),  face and hands shape $\beta \in \mathbb{R}^{|\beta|}$ and facial expression $\psi \in  \mathbb{R}^{|\psi|}$. To be specific:
\begin{equation}\label{eq4}
    M(\beta, \theta, \psi)=W\left(T_p(\beta, \theta, \psi), \mathcal{J}(\beta), \theta, \mathcal{W}\right),
\end{equation}
where $T_p$ is the human body mesh in the canonical pose, $\mathcal{J}$ is the pre-trained regression matrix, $W$ is the pose transformation operation, and $\mathcal{W}$ is the predetermined skin blending weight. For more details refer to~\cite{SMPL-X:2019}.


\section{Proposed Method}
We present the pipeline of our proposed method in Figure~\ref{fig3}, which comprises three key components: (1) Drivable avatar representation based on 3D GS (Sec.~\ref{sec41}), (2) Pose refinement for avatar reconstruction (Sec.~\ref{sec42}), and (3) Surface-guided Gaussian re-initialization (Sec.~\ref{sec43}).


\begin{figure*}[t]
\centering
\includegraphics[width=\linewidth]{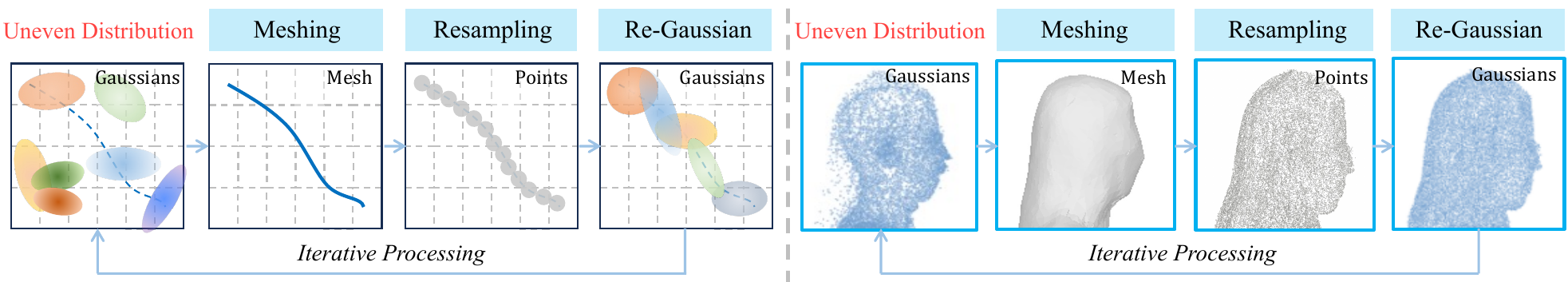}
\vspace{-0.5cm}
\caption{The surface-guided re-initialization mechanism uses the three operations of \textit{Meshing, Resampling, and Re-Gaussian} to redistribute unevenly Gaussian points near the real surface, thereby enhancing the stability of the avatar in novel poses.}
\label{fig4}
\end{figure*}

\subsection{The Representation of 3D Gaussian Avatars}\label{sec41}
Our 3D Gaussian avatar model comprises two key components represented as $\{G, B\}$. The first is a collection of 3D Gaussian points, denoted as $G$, which captures the target subject's shape and appearance characteristics. The second is a comprehensive skeleton model, represented as $B$, which allows for avatar manipulation.

We initialize the Gaussian points $G$ in the canonical pose space (i.e., T-pose) by utilizing the vertices of the SMPL-X model.
To enable deformations and pose variations in our avatars, we utilize the SMPL-X bone structure~\cite{SMPL-X:2019} for the skeleton. This skeleton consists of $K=55$ joints, with $22$ joints responsible for controlling the body pose, $15\times2$ joints controlling the left and right hands, respectively, and the remaining $3$ joints controlling the head. We employ the joint hierarchy to achieve pose transformations and calculate the pose transformation matrix $\mathcal{T}(\theta)$ for each joint, given a pose $\theta$. For each Gaussian point, we calculate the pose transformation $\mathcal{A}$ based on the nearest $P=4$ joints through the following formula:
\begin{equation}\label{eq5}
    \mathcal{A}(\theta) = \sum_{p=1}^{P}\mathcal{W}_p(\bm \mu)\mathcal{T}(\theta),
\end{equation}
where $\mathcal{W}_p(\bm  \mu)$ is the skinning of the Gaussian point $\bm \mu$, obtained by finding the skinning of the nearest neighbor vertex of SMPL-X. The deformation of the Gaussian point from the canonical pose to the target pose $\theta$ can be written as:
\begin{equation}\label{eq6}
   \bm  \mu_{\theta} = \mathcal{A}_\text{rot}(\theta)\bm \mu'+ \mathcal{A}_{t}, \quad \bm R_{\theta} = \mathcal{A}_\text{rot}(\theta)\bm R,
\end{equation}
where $\mathcal{A}_\text{rot}$ represents the rotation component, and $\mathcal{A}_{t}$ represents the translation component of the Gaussian point transformation. $\bm R$ is the rotation matrix of the Gaussian point calculated from its quaternion $\vq$. To address non-rigid local deformations, such as those occurring in garments, we introduce an adjusted Gaussian position $\bm \mu'$. This adjustment is achieved by adding the pose-conditioned residual to the original Gaussian position written as $\bm \mu' = \bm \mu + \text{MLP}(\theta)$.



\subsection{Pose Refinement for Avatar Reconstruction}\label{sec42}

Creating a high-quality 3D Gaussian avatar hinges significantly on the precision of pose estimation derived from the input images. This dependency stems from accurate pose data being crucial for properly aligning the 3D Gaussian avatar with the captured images.
Limited by the current whole-body pose estimation method's~\cite{zhang2023pymaf,lin2023one,li2023hybrik} inability to align the hand and foot areas stably, the exiting 3D Gaussian-based avatar method~\cite{zielonka2023drivable,yuan2023gavatar,qian20233dgs,saito2023relightable} still only focuses on body-controllable reconstruction and does not support finer-grained hand control.

To tackle this challenge, we introduce a two-stage method that specifically focuses on enhancing the accuracy of whole-body poses. 
Concretely, in the first stage, we obtain an initial pose estimation by using an existing whole-body pose estimation network $\mathcal{E}$~\cite{zhang2023pymaf} applied to the frame data $I$ from the given video. This process allows us to derive the SMPL-X pose parameters $\theta$ and camera parameters $\Pi$ as the coarse whole-body pose estimation result.
\begin{equation}\label{eq8}
    \theta^{\text{stage1}}, \Pi = \mathcal{E}(I).
\end{equation}
The poses obtained from this stage often exhibit noticeable misalignment when the human subject is positioned sideways, especially in the hand and foot regions, as depicted in Figure~\ref{fig11}.

In the second stage, we incorporate constraints from normal maps and silhouettes to optimize the pose further, aiming for enhanced congruency between the SMPL-X model and the subjects depicted in the images.  The critical insight is that 1) the normal map can effectively guide the alignment of the whole body, especially hand pose and feet, and 2) silhouettes act as a boundary condition, guaranteeing that the hand and foot areas precisely match the actual placements observed in the images. Specifically, for a given input image, we use Segment Anything Model (SAM)~\cite{kirillov2023segment} to obtain the mask of the target subject as its silhouette $S^{\text{pred}}$, and then use ICON~\cite{xiu2022icon} to obtain predicted normal $N^{\text{pred}}$. This loss function is as follows: 
\begin{equation}\label{eq9}\footnotesize
    \mathcal{L}_\text{pose} = \underbrace{\left|N - N^\text{pred}\right|}_{\mathcal{L}_\text{normal}} + \lambda_{1}\underbrace{\left|S - S^\text{pred}\right|}_{\mathcal{L}_\text{silhouette}} + \\ \lambda_{2}\underbrace{\sum_{i=1}^{K}\omega_{i}\left(\theta_{i} - \theta_{i}^\text{stage1}\right)}_{\mathcal{L}_\text{regular}},
\end{equation}
where $\lambda$ is used to weight different loss terms. In the experiment, we empirically set $\lambda_{1} = 5.0 $ and $\lambda_{2} = 0.5$.
The loss function consists of three terms. The first term enforces consistency between the rendered normal map $N$ from SMPL-X using current pose parameters $\theta$ and the predicted normal map $N^\text{pred}$ from the image. The second term ensures alignment between the rendered silhouette and the predicted silhouette of the subject. The third term regularizes the optimizing pose $\theta$ to remain close to the estimated significant pose $\theta^\text{stage1}$ from the first stage. We apply a weighting mechanism to different joints based on their distance from the root joint, assigning lower weights to joints further away.

\begin{figure*}[t]
\centering
\includegraphics[scale=0.6]{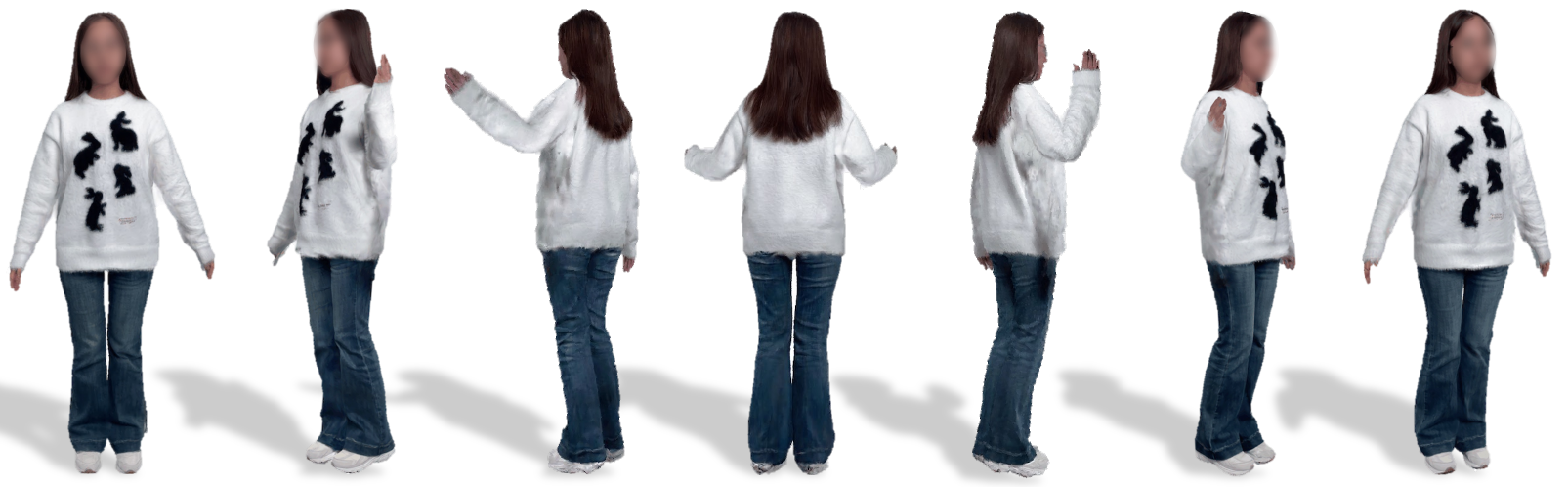}
\vspace{-0.2cm}
\caption{Rendered frames of our reconstructed Gaussian avatar from novel views.}
\label{fig5}
\end{figure*}

\begin{figure*}[t]
\centering
\includegraphics[scale=0.53]{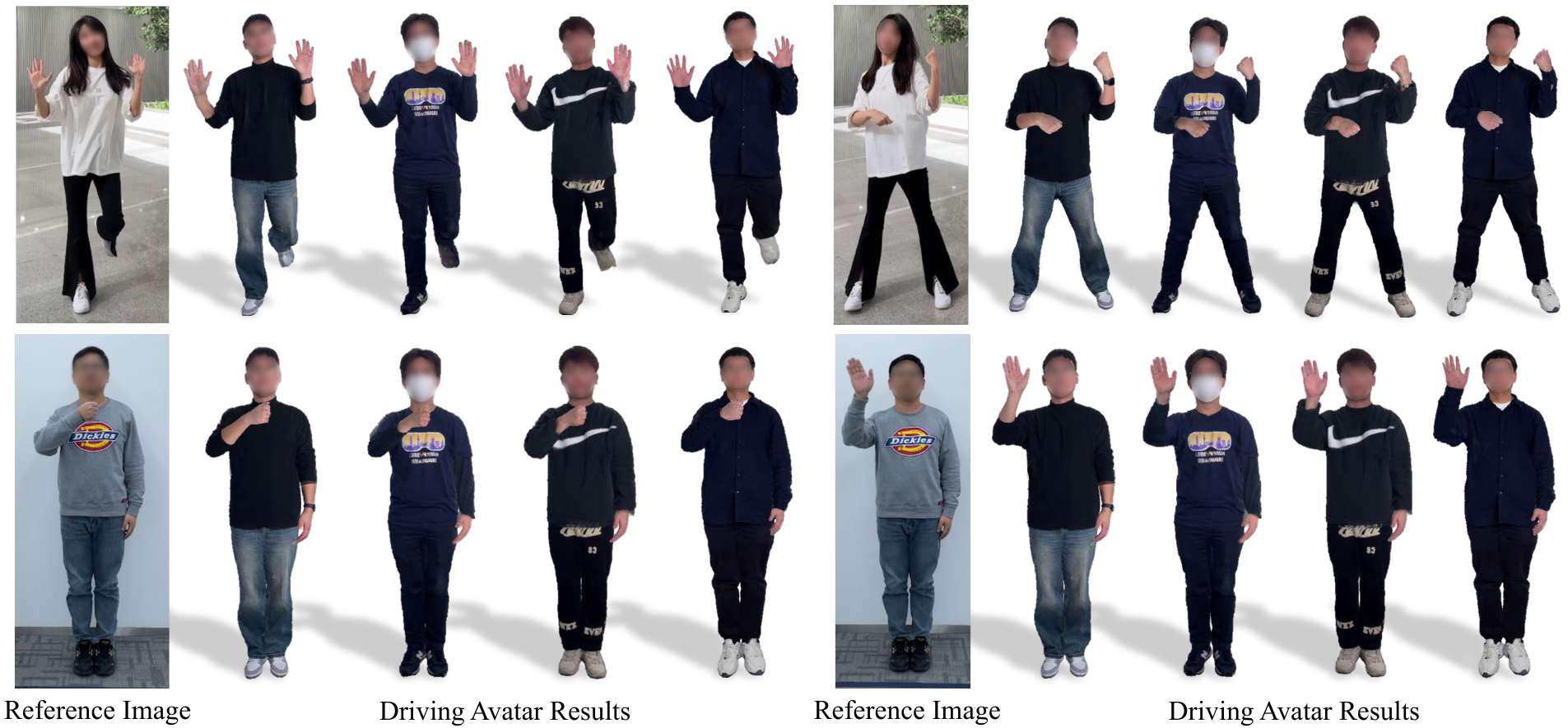}
\caption{Multiple reconstructed avatars demonstrate pose-driven movements using videos collected from real-world.} 
\label{fig6}
\end{figure*}

\subsection{Surface-Guided Gaussian Re-Initialization}\label{sec43}
This section introduces a surface-guided Gaussian re-initialization method to tackle unbalanced aggregation and initialization bias that degrade the performance of 3D Gaussian avatars. 
The unbalanced aggregation results from the cloning and splitting operations of 3DGS, which propagate Gaussian points in high-frequency texture areas, resulting in local aggregation. Meanwhile, 3D Gaussian points are susceptible to initialization, which further exacerbates the artifacts in the avatar model. 

Existing 3D Gaussian avatars usually use the SMPL~\cite{loper2023smpl,SMPL-X:2019} to initialize 3D Gaussian points. 
While 3D Gaussian representation is viable for subjects with tight clothes, reconstructing subjects with long-haired shawls or loose garments still poses challenges. In such cases, 3D Gaussian points tend to spread outside the human body. Consequently, when these regions undergo significant pose deformations, those falsely distributed Gaussian points often result in blurriness and artifacts in the rendered frames.



\begin{figure*}[t]
\centering
\includegraphics[width=1.0\linewidth]{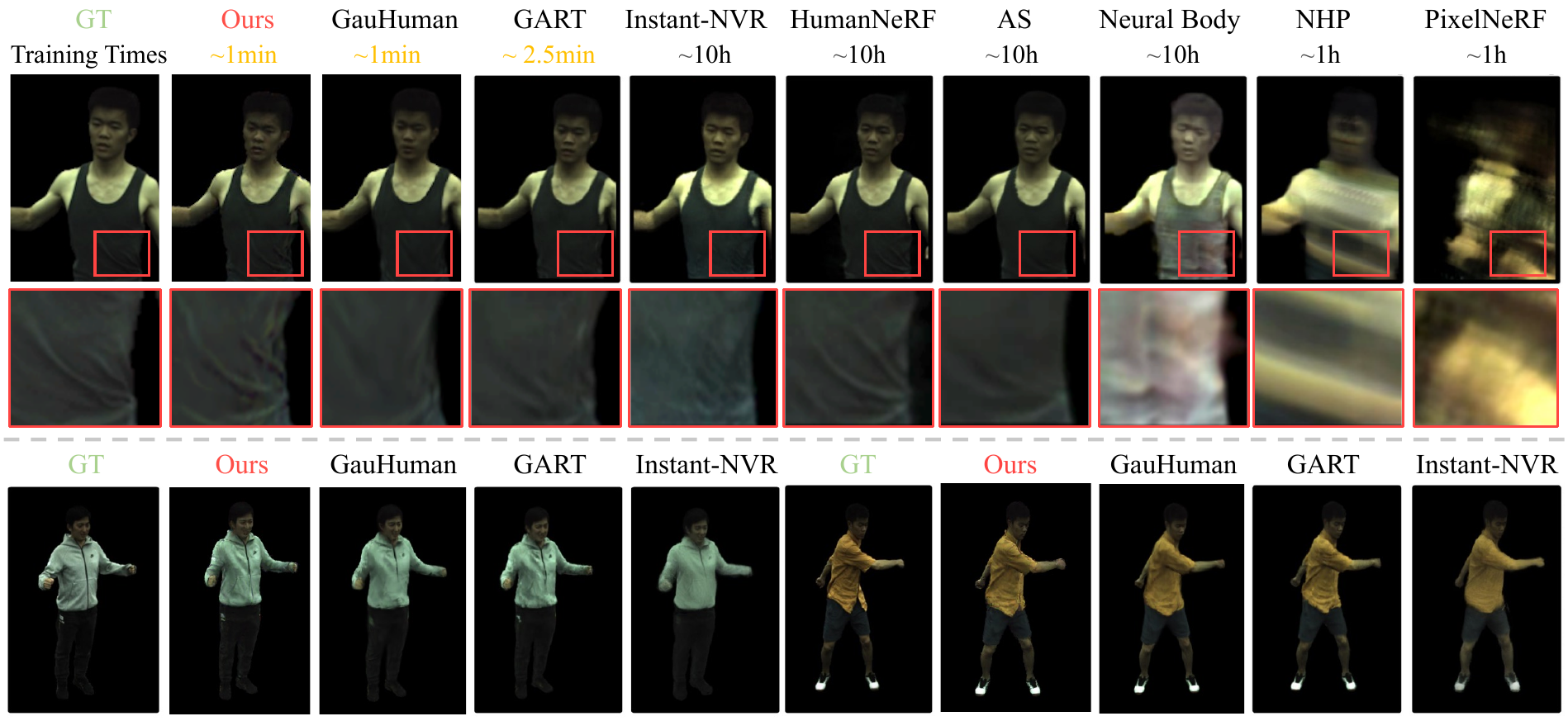}
\vspace{-0.5cm}
\caption{Qualitative comparison on the ZJU-MoCap \cite{peng2021neural} dataset.}
\label{fig7}
\end{figure*}

\begin{table*}[t]
\centering
\fontsize{8pt}{10pt}\selectfont
\setlength{\tabcolsep}{26pt}
\caption{Quantitative comparison on the ZJU-MoCap~\cite{peng2021neural} dataset. \textcolor[rgb]{ 1, 0.7, 0.7}{\textbf{Pink}} highlights the best, and \textcolor[rgb]{1, 0.85, 0.7}{\textbf{orange}} highlights the second best.}
\vspace{0.5cm}
\begin{tabular}{lcccc}
\toprule[1.2pt]
Methods                                                            & PSNR$\uparrow$                                                 & SSIM$\uparrow$                                                  & LPIPS$^*$$\downarrow$                                          & Training time                                                                   \\ \hline
HumanNeRF~\cite{weng2022humannerf} & 30.66                                                          & 0.9690                                                          & 33.38                                                          & $\sim$ 10 h                                                             \\
AS~\cite{peng2024animatable}                 & 30.38                                                          & 0.9750                                                          & 37.23                                                          & $\sim$ 10 h                                                             \\
AN~\cite{peng2021animatable}                 & 29.77                                                          & 0.9652                                                          & 46.89                                                          & $\sim$ 10 h                                                             \\
Neural Body~\cite{peng2021neural}            & 29.03                                                          & 0.9641                                                          & 42.47                                                          & $\sim$ 10 h                                                             \\
DVA~\cite{remelli2022drivable}               & 29.45                                                          & 0.9564                                                          & 37.74                                                          & $\sim$ 1.5 h                                                            \\ \hline
NHP~\cite{kwon2021neural}                    & 28.25                                                          & 0.9551                                                          & 64.77                                                          & $\sim$ 1 h tuning                                                       \\
PixelNeRF~\cite{yu2021pixelNeRF}             & 24.71                                                          & 0.8920                                                          & 121.86                                                         & $\sim$ 1 h tuning                                                       \\ \hline
Instant-NVR~\cite{instant_nvr}              & 31.01                                                          & 0.9710                                                          & 38.45                                                          & $\sim$ 5 min                                                            \\
Instant-Avatar~\cite{jiang2023instantavatar} & 29.73                                                          & 0.9384                                                          & 68.41                                                          & $\sim$ 3 min                                                            \\ \hline
GauHuman~\cite{hu2023gauhuman}               & 31.34                                                          & 0.9650                                                          & 30.51                                                          & \cellcolor[rgb]{ 1, 0.7, 0.7}$\sim$ 1 min   \\
GART~\cite{lei2023gart}                      & \cellcolor[rgb]{1, 0.85, 0.7}32.22 & \cellcolor[rgb]{1, 0.85, 0.7}0.9771 & \cellcolor[rgb]{1, 0.85, 0.7}29.21 & \cellcolor[rgb]{1, 0.85, 0.7}$\sim$ 2.5 min \\
\textbf{Ours}                                     & \cellcolor[rgb]{ 1, 0.7, 0.7}32.45 & \cellcolor[rgb]{ 1, 0.7, 0.7}0.9773 & \cellcolor[rgb]{ 1, 0.7, 0.7}26.94 & \cellcolor[rgb]{ 1, 0.7, 0.7}$\sim$ 1 min   \\ \bottomrule[1.2pt]
\label{tab1}
\end{tabular}
\vspace{-0.4cm}
\end{table*}

Our key insight is to impose additional constraints on the distribution of 3D Gaussian points, enforcing them to be uniformly distributed near the subject's surface. To this end, we propose a surface-guided Gaussian re-initialization method, as shown in Figure~\ref{fig4}, which includes three operations to be iteratively applied to the Gaussian avatars: \textit{Meshing, Resampling, and Re-Gaussian}. Meshing provides geometric surface priors as a constraint for 3D Gaussian points, resampling is utilized to constrain the 3D Gaussian distribution to be uniform, and Re-Gaussian is used to avoid falling into a local minimum state. We iteratively perform the proposed mechanism for 2-3 times so that the Gaussian avatar can gradually approach the real surface of the human body.

\textbf{Meshing.} We use the spherical shell surface reconstruction method~\cite{edelsbrunner1983shape} to obtain the avatar's surface mesh, represented by the human body's outermost Gaussian points.

\textbf{Resampling.} We perform Laplacian smoothing for the reconstructed avatar mesh to inject a surface smoothness prior. Then, We carry out curvature-based uniform sampling on the mesh as new Gaussian points.

\textbf{Re-Gaussian.} We find their K-nearest Gaussian points for the resampled points and inherit their opacity $\eta$ and spherical coefficient $\vf$ properties. The rotation $\bm R$ and scaling $\vs$ properties are randomly reinitialized.

\subsection{Differentiable Rendering Loss Function}\label{sec44}
We use SMPL-X skeleton transformation (Eq.~\ref{eq5} and Eq.~\ref{eq6}) to drive the Gaussian avatar from the canonical space to the image space and optimize it with differentiable rendering.
Given the rendered image $C$ and the input image $I$, we calculate the reconstruction loss $\mathcal{L}_\text{recon}$, perceptual loss $\mathcal{L}_\text{perceptual}$, and residual regularization $\mathcal{L}_\text{residual}$. The total loss function is
\begin{equation}\label{eq13}\footnotesize
    \mathcal{L}_\text{render} = \underbrace{\left|C - I\right|}_{\mathcal{L}_\text{recon}} + \lambda_{3}\underbrace{\left|\text{VGG}(C) - \text{VGG}(I)\right|}_{ \mathcal{L}_\text{perceptual}} + \lambda_{4}\underbrace{\left|\text{MLP}(\theta)\right|}_{\mathcal{L}_\text{residual}}.
\end{equation}
We empirically set $\lambda_{3} = 0.1$ and $\lambda_{4} = 0.5$.
The reconstruction term $\mathcal{L}_\text{recon}$ constrains the rendered avatar image $C$ to be consistent with the input image $I$. The perceptual term loss $\mathcal{L}_{perceptual}$ constrains the rendered image $C$ and the input image $I$ to have consistent encoded features, which can ensure the effective learning of high-frequency appearance details. $\text{VGG}(*)$ represents the high-dimensional image features from the pre-trained VGG network~\cite{SimonyanZ14a}. The residual regularization $\mathcal{L}_{residual}$ is used to regularize the pose-conditioned residual to zero to avoid significantly interfering with the Gaussian avatar.

%% file: 4_experiments.tex

\section{Experiments}


\begin{figure*}[t]
\centering
\includegraphics[width=1.0\linewidth]{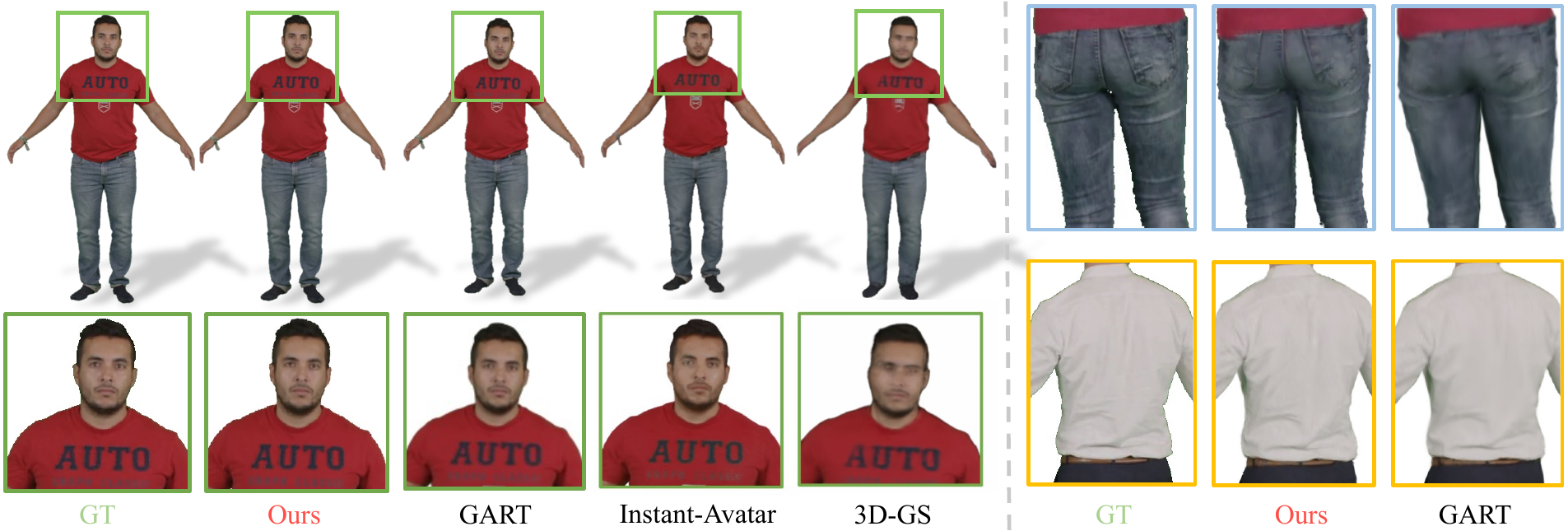}
\vspace{-0.4cm}
\caption{Qualitative evaluation on the People-Snapshot~\cite{alldieck2018video} dataset, comparing our method with multiple baseline approaches.}
\vspace{-0.3cm}
\label{fig8}
\end{figure*}

\begin{table*}[t]
\fontsize{8pt}{12pt}\selectfont
\setlength{\tabcolsep}{4.5pt}
  \centering
  \caption{Quantitative comparison on the People-Snapshot~\cite{alldieck2018video} dataset.}
  \vspace{15pt}
    \begin{tabular}{lcccccccccccc}
    \toprule[1.2pt]
    \multirow{2}[2]{*}{Methods} & \multicolumn{3}{c}{male-3-casual} & \multicolumn{3}{c}{male-4-casual} & \multicolumn{3}{c}{female-3-casual} & \multicolumn{3}{c}{female-4-casual} \\
          & PSNR  & SSIM  & LPIPS & PSNR  & SSIM  & LPIPS & PSNR  & SSIM  & LPIPS & PSNR  & SSIM  & LPIPS \\
    \midrule
    3D-GS~\cite{kerbl20233d} & 26.60  & 0.9393 & 0.0820 & 24.54 & 0.9469 & 0.0880 & 24.73 & 0.9297 & 0.0930 & 25.74 & 0.9364 & 0.0750 \\
    Neural Body~\cite{peng2021neural} & 24.94  & 0.9428 & 0.0326 & 24.71 & 0.9469 & 0.0423 & 23.87 & 0.9504 & 0.0346 & 24.37 & 0.9451 & 0.0382 \\
    Anim-NeRF~\cite{chen2021animatable} & 12.39  & 0.7929 & 0.3393 & 13.10 & 0.7705 & 0.3460 & 11.71 & 0.7797 & 0.3321 & 12.31 & 0.8089 & 0.3344 \\
    Instant-Avatar~\cite{jiang2023instantavatar} & 29.65  & 0.9730 & \cellcolor[rgb]{ 1, 0.7, 0.7}0.0192 & \cellcolor[rgb]{ 1, 0.7, 0.7}27.97 & 0.9649 & \cellcolor[rgb]{ 1, 0.7, 0.7}0.0346 & \cellcolor[rgb]{ 1, 0.7, 0.7}27.90 & 0.9722 & \cellcolor[rgb]{ 1, 0.7, 0.7}0.0249 & 28.92 & 0.9692 & \cellcolor[rgb]{ 1, 0.7, 0.7}0.0180 \\
    GART~\cite{lei2023gart}  & \cellcolor[rgb]{1, 0.85, 0.7}30.40  & \cellcolor[rgb]{1, 0.85, 0.7}0.9769 & 0.0377 & 27.57 & \cellcolor[rgb]{1, 0.85, 0.7}0.9657 & 0.0607 & 26.26 & \cellcolor[rgb]{1, 0.85, 0.7}0.9656 & 0.0498 & \cellcolor[rgb]{1, 0.85, 0.7}29.23 & \cellcolor[rgb]{1, 0.85, 0.7}0.9721 & 0.0378 \\
    \textbf{Ours}  & \cellcolor[rgb]{ 1, 0.7, 0.7}30.82  & \cellcolor[rgb]{ 1, 0.7, 0.7}0.9808 & \cellcolor[rgb]{1, 0.85, 0.7}0.0199 & \cellcolor[rgb]{1, 0.85, 0.7}27.62 & \cellcolor[rgb]{ 1, 0.7, 0.7}0.9742 & \cellcolor[rgb]{1, 0.85, 0.7}0.0351 & 25.93 & \cellcolor[rgb]{ 1, 0.7, 0.7}0.9684 & \cellcolor[rgb]{1, 0.85, 0.7}0.0325 & \cellcolor[rgb]{ 1, 0.7, 0.7}29.27 & \cellcolor[rgb]{ 1, 0.7, 0.7}0.9743 & \cellcolor[rgb]{1, 0.85, 0.7}0.0213 \\
    \bottomrule[1.2pt]
    \end{tabular}%

  \label{tab2}%
\end{table*}%

\subsection{Setup and Datasets}

Our approach is based on the PyTorch framework and utilizes the Adam optimizer. The model is optimized for $3,000$ steps, with the learning rate for the Gaussian's position, rotation, scale, transparency, and spherical harmonic coefficient all set similarly to \cite{lei2023gart}. The experiment is conducted on an NVIDIA A100 GPU, with pose refinement requiring 10 seconds per frame.



\textbf{People-Snapshot}~\cite{alldieck2018video} is a monocular video dataset, which contains 8 subjects wearing various clothing and performing self-rotation motions in front of a fixed camera, maintaining an A-pose during the recording. 

\textbf{ZJU-MoCap}~\cite{peng2021neural} is a multi-view dataset that includes dynamic videos of 6 subjects captured by over 20 simultaneous cameras. 

ZJU-MoCap and People-Snapshot lack diversity in hand pose, therefore, we introduce the GVA-Snapshot dataset.



\textbf{GVA-Snapshot} dataset is intended for evaluating body and hand reconstruction from monocular videos. It includes self-rotation videos and carefully designed hand movement videos of 7 subjects. Each data frame provides 4K resolution RGB images, precise masks, and corresponding refined SMPL-X pose parameters. Additionally, our subjects exhibit challenging features such as shawl-length hair, which are absent in current public datasets. More details are presented in the supplementary materials.




\subsection{Baselines and Evaluation Metrics}

Baseline methods can be categorized into NeRF-based and 3D Gaussian-based approaches, based on the avatar representation. NeRF-based methods such as HumanNeRF~\cite{weng2022humannerf}, AS~\cite{peng2024animatable}, AN~\cite{peng2021animatable}, Neural Body~\cite{peng2021neural}, DVA~\cite{remelli2022drivable}, NHP~\cite{kwon2021neural}, PixelNeRF~\cite{yu2021pixelNeRF}, Instant-NVR~\cite{instant_nvr}, and Instant-Avatar~\cite{jiang2023instantavatar} employ different variations of the NeRF representation for avatar reconstruction. HumanNeRF, AS, AN, Neural Body, and DVA utilize a naive NeRF representation combined with locally encoded human body features. NHP and PixelNeRF use a generalizable NeRF representation, reducing training time through finetuning. Instant-NVR and Instant-Avatar enable NeRF representation for minute-level training and real-time rendering using grid hashing. Gaussian-based methods, including GauHuman~\cite{hu2023gauhuman} and GART~\cite{lei2023gart}, represent the current state-of-the-art approaches for Gaussian avatar reconstruction.


For quantitative evaluation, we use three metrics: Peak Signal-to-Noise Ratio (PSNR), Structural Similarity Index Measure (SSIM), and Learned Perceptual Image Patch Similarity (LPIPS)~\cite{zhang2018perceptual}. PSNR is used to evaluate pixel-level errors between avatar-rendered images and ground-truth images. SSIM is used to assess structure-level errors, while LPIPS evaluates perceptual errors.

\begin{figure}[t]
\centering
\includegraphics[scale=0.64]{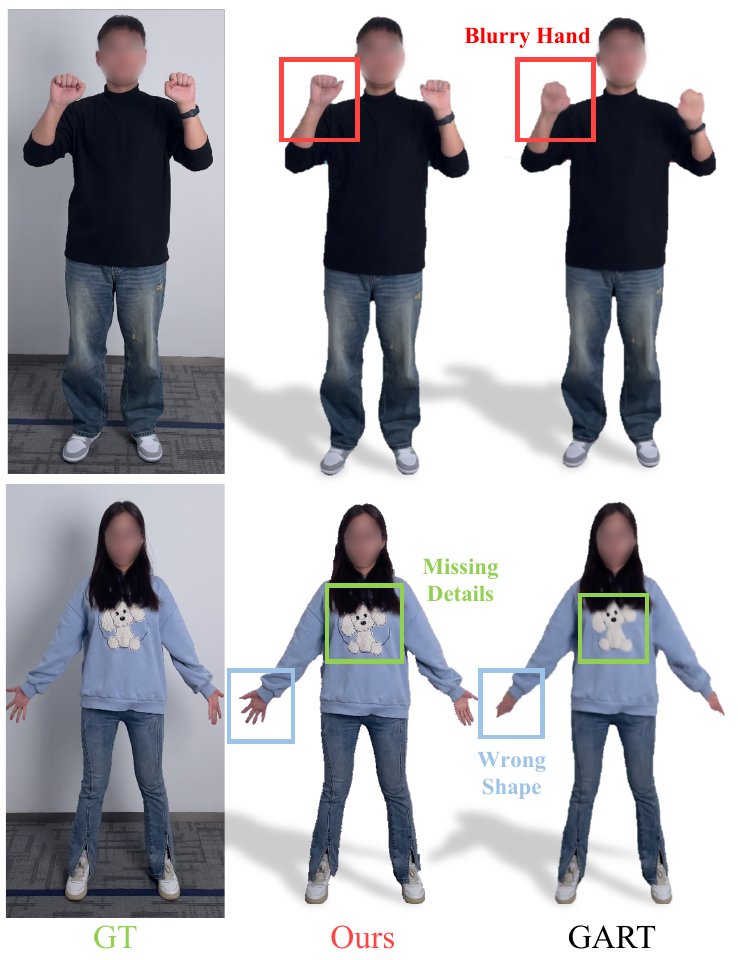}
\caption{Comparison of results between ours and GART~\cite{lei2023gart} on GVA-Snapshot dataset.} 
\label{fig9}
\end{figure}

\begin{figure}[t]
\centering
\includegraphics[scale=0.43]{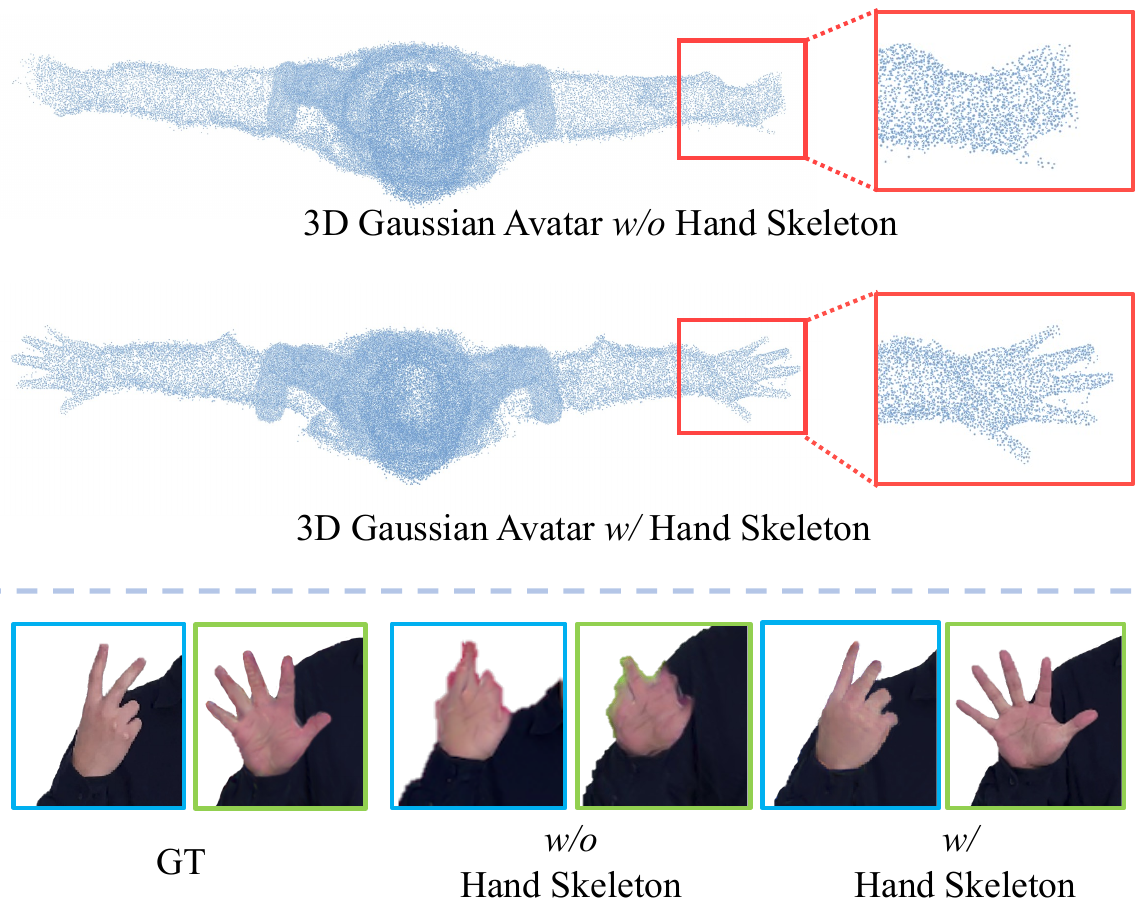}
\vspace{-0.1cm}
\caption{Ablation study investigating the impact of the hand skeleton. Top: 3D Gaussian avatar visualization; Bottom: Zoom-in rendered images.}
\label{fig10}
\end{figure}

\subsection{Qualitative Experiments}
Three qualitative experiments are conducted to demonstrate the effectiveness of our proposed method as follows.

First, we showcase the capability of our method to render reconstructed avatars from various novel viewpoints, as shown in Figure~\ref{fig5}. This demonstrates the ability to reconstruct complete and visually accurate avatar models from monocular videos, capturing photorealistic effects from different perspectives. Additionally, we utilize a video captured in natural settings to estimate its SMPL-X pose as a driving sequence, enabling whole-body pose control and motion reproduction for the avatar, as depicted in Figure~\ref{fig6}. Our reconstructed avatar maintains fidelity in details and accurately represents hand movements when driven to unseen poses, highlighting the strong generalization ability.

Second, we evaluate our method against multiple baseline methods on the ZJU-MoCap and People-Snapshot~\cite{alldieck2018video} dataset, as shown in Figure~\ref{fig7} and Figure~\ref{fig8}. Compared to AS~\cite{peng2024animatable}, NB~\cite{peng2021neural}, NHP~\cite{kwon2021neural}, PixelNeRF~\cite{yu2021pixelNeRF}, and Instant-NVR~\cite{instant_nvr}, our method demonstrates superior accuracy in capturing shape and appearance from novel views. Compared to HumanNeRF~\cite{weng2022humannerf} in Figure~\ref{fig7}, our method achieves a visually comparable performance with significantly reduced time consumption. Compared to GART~\cite{lei2023gart} and Instant-Avatar~\cite{jiang2023instantavatar} in Figure~\ref{fig8}, our method captures more details. These results highlight our method's advantages in realism and efficiency.

Third, we compare our approach with GART~\cite{lei2023gart} on the GVA-Snapshot dataset, as depicted in Figure~\ref{fig9}. GART\cite{lei2023gart}, which uses SMPL as the skeleton without hand pose guidance, shows incorrect shapes and blurred hands. In contrast, our method incorporates the SMPL-X skeleton and incorporates hand guidance, enabling full-body pose control for the avatar and providing more precise details.

\begin{figure*}[tbp]
\centering
\includegraphics[width=0.99\linewidth]{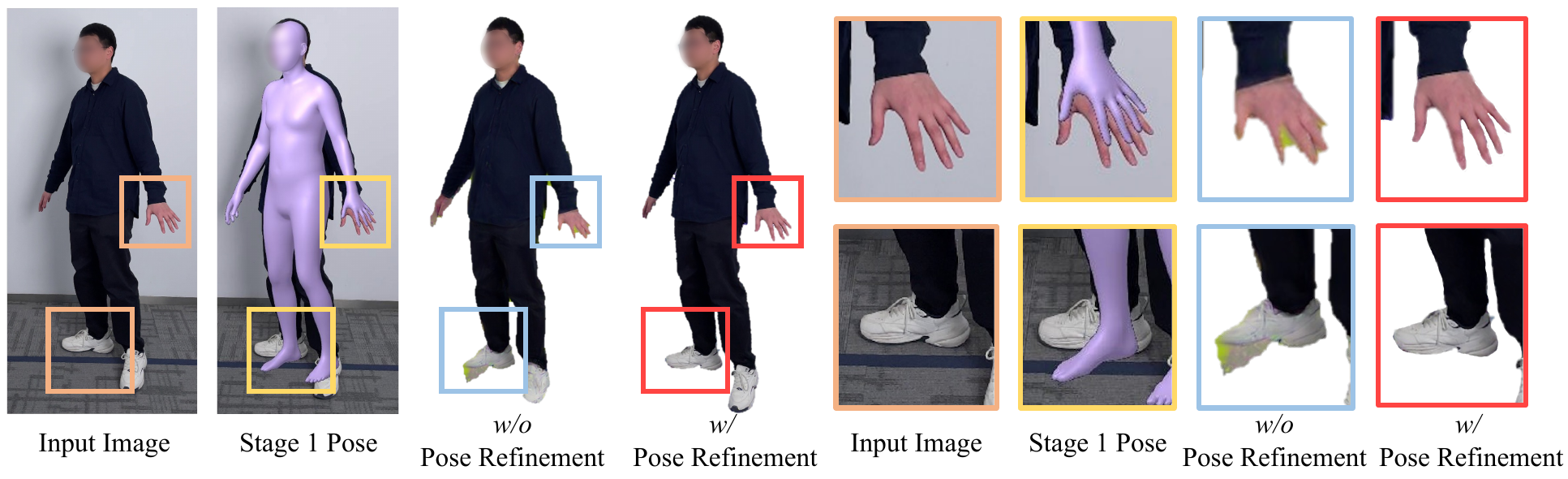}
\vspace{-0.1cm}
\caption{Ablation study on utilizing the pose refinement approach.}
\vspace{-0.0cm}
\label{fig11}
\end{figure*}

\begin{figure*}[t]
\centering
\includegraphics[width=0.99\linewidth]{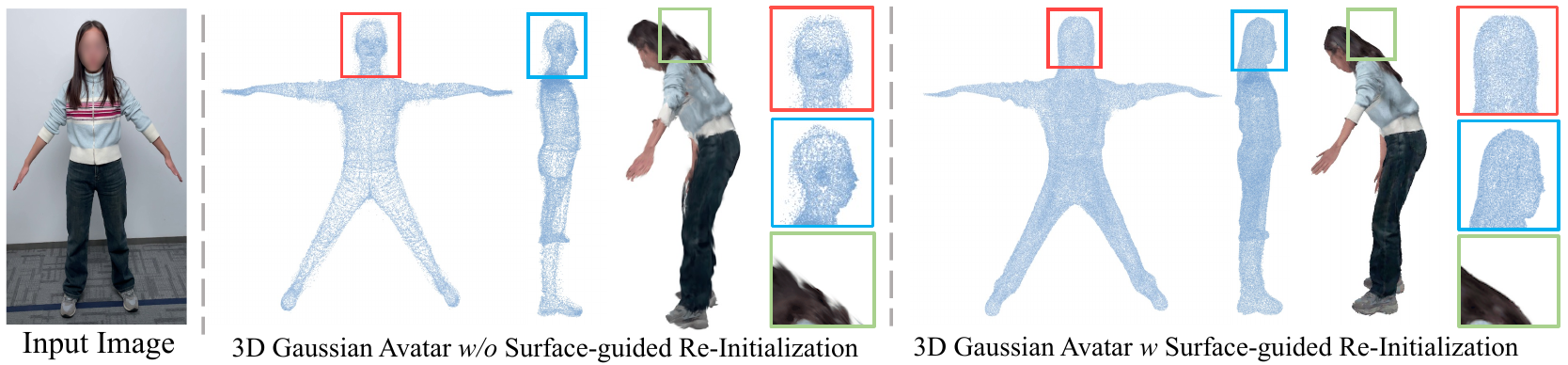}
\caption{Ablation study on the utilization of the surface-guided re-initialization.}
\vspace{-0.2cm}
\label{fig12}
\end{figure*}

\subsection{Quantitative Results}


In Tabel~\ref{tab1} and Tabel~\ref{tab2}, we compare our method with baseline methods on the ZJU-MoCap and Peopel-Snapshot datasets. Our method notably outperforms various NeRF-based methods, is on par with GART~\cite{lei2023gart} in terms of PSNR and SSIM, and significantly outperforms in LPIPS. These results align with qualitative observations.
Given the absence of hand pose changes in the above two datasets, we compare our method with GART~\cite{lei2023gart} on the GVA-Snapshot dataset. 
The comparative results are detailed in Table~\ref{tab3}. These results indicate that our method outperforms GART~\cite{lei2023gart} across all metrics, consistent with the qualitative assessment. These observations indicate that our method attains superior avatar reconstruction performance.

\begin{table}[t]
\renewcommand\arraystretch{1.3}
\fontsize{8pt}{11pt}\selectfont
  \centering
  \tabcolsep=0.34cm
  \caption{Quantitative comparison between ours and GART~\cite{lei2023gart} on GVA-Snapshot.}
  \vspace{15pt}
  \begin{tabular}{lccc}
    \toprule[1.0pt]
    Methods & PSNR$\uparrow$  & SSIM$\uparrow$  &   LPIPS$^*$$\downarrow$ \\
    \midrule
    GART~\cite{lei2023gart}  & 31.61 & 0.9907 & 38.52 \\
    \textbf{Ours}  & \cellcolor[rgb]{1, 0.7, 0.7}32.36 & \cellcolor[rgb]{1, 0.7, 0.7}0.9912 & \cellcolor[rgb]{1, 0.7, 0.7}27.24 \\
    \bottomrule[1.0pt]
\end{tabular} 
\label{tab3}
\end{table}%

\subsection{Ablation Study}





This section examines the influence of key technical components, namely the addition of hand skeleton, pose refinement, and surface-guided re-initialization. 

Figure~\ref{fig10} illustrates the effect of incorporating a hand skeleton on the reconstructed avatar. Without the hand skeleton, Gaussian points struggle to capture the hand shape accurately, leading to blurred images.
Figure~\ref{fig11} explores the influence of employing pose refinement. The comparison primarily focuses on the avatar results obtained through solely one-stage pose estimation. The findings reveal that relying solely on the existing whole-body pose estimation (without pose refinement) fails to completely align the subject's pose in the image, particularly in sideways situations. This inadequacy leads to significant artifacts in the foot region of the learned avatar. However, with increased pose refinement, the avatar acquires more accurate pose guidance, effectively mitigating this issue.


\begin{table}[t]
\renewcommand\arraystretch{1.0}
\fontsize{8pt}{11pt}\selectfont
  \centering  
  \tabcolsep=0.18cm
  \caption{Quantitative ablation study on different components.}
  \vspace{15pt}
  \begin{tabular}{lccc}
    \toprule[1.0pt]
    Methods & PSNR$\uparrow$  & SSIM$\uparrow$  &  LPIPS$^*$$\downarrow$ \\
    \midrule
    \textit{w/o} Pose Refinement & 26.76 & 0.978 & 51.20 \\
    \textit{w/o} Hand Skeleton & 28.79 & 0.982 & 34.45 \\
    \textit{w/o} Surface-guided Re-Initialization & 30.48 & \cellcolor[rgb]{1, 0.7, 0.7}0.989 & 35.30 \\
    \textbf{Ours (Full)} & \cellcolor[rgb]{1, 0.7, 0.7}32.22 & \cellcolor[rgb]{1, 0.7, 0.7}0.989 & \cellcolor[rgb]{1, 0.7, 0.7}31.80 \\
    \bottomrule[1.0pt]
    \end{tabular}%
    \label{tab4}
\label{tab3}
\end{table}%




Figure~\ref{fig12} illustrates the impact of employing surface-guided re-initialization. Without surface-guided re-initialization, Gaussian points are only sparsely allocated in the external areas of the naked body (such as hair), making the avatar susceptible to noticeable artifacts undergoing new pose drives. Conversely, utilizing surface-guided re-initialization effectively redistributes the avatar's Gaussian points, ensuring a more even distribution across the real human body surface, thus enhancing the stability of new pose results.


Table~\ref{tab4} illustrates the quantitative ablation results. In alignment with the qualitative analysis, it demonstrates that each technical component contributes positively to the final body-hand avatar reconstruction results.




%% file: 5_conclusion.tex
\section{Conclusion}
This paper proposes a body- and hand-drivable 3D Gaussian avatar reconstruction method from monocular videos. This method utilizes a pose refinement to improve hand and foot pose accuracy, thereby guiding Avatar to learn the correct shape and appearance. Furthermore, a surface-guided Gaussian re-initialization mechanism is introduced to alleviate the unbalanced aggregation and initialization bias problems. Our aim is that this contribution will pave the way for more lifelike avatar reconstructions in future endeavors.

\vspace{4pt}
\noindent \textbf{Limitation.} Although our method has successfully achieved body- and hand-controllable avatar reconstruction, further increasing facial expression controllability remains a challenge. Introducing learnable blendshapes may be a feasible way. In addition, our method is currently unable to directly handle very loose clothing, such as long skirts. Introducing physical-based deformation priors may be a worthwhile approach, such as~\cite{xie2023physgaussian}, to explore it in the future.

\vspace{4pt}
\noindent \textbf{Potential Negative Impact.} Our methods may invade privacy or be used by criminals for improper purposes. Therefore, watermarking technology and related regulations need to be improved to ensure that the technology can be used safely and serve society.

\section{Acknowledgments}
The authors would like to thank Xiaobo Gao, Chunyu Song, Yanmin Wu, Hao Li, Lingyun Wang, Zhenxiong Ren, and Haotian Peng for helping.




%% file: main.bbl
\begin{thebibliography}{60}
\providecommand{\natexlab}[1]{#1}
\providecommand{\url}[1]{\texttt{#1}}
\expandafter\ifx\csname urlstyle\endcsname\relax
  \providecommand{\doi}[1]{doi: #1}\else
  \providecommand{\doi}{doi: \begingroup \urlstyle{rm}\Url}\fi

\bibitem[Alldieck et~al.(2018)Alldieck, Magnor, Xu, Theobalt, and
  Pons-Moll]{alldieck2018video}
Alldieck, T., Magnor, M., Xu, W., Theobalt, C., and Pons-Moll, G.
\newblock Video based reconstruction of 3d people models.
\newblock In \emph{CVPR}, pp.\  8387--8397, 2018.

\bibitem[Barron et~al.(2021)Barron, Mildenhall, Tancik, Hedman, Martin-Brualla,
  and Srinivasan]{barron2021mip}
Barron, J.~T., Mildenhall, B., Tancik, M., Hedman, P., Martin-Brualla, R., and
  Srinivasan, P.~P.
\newblock Mip-nerf: A multiscale representation for anti-aliasing neural
  radiance fields.
\newblock In \emph{ICCV}, pp.\  5855--5864, 2021.

\bibitem[Chen et~al.(2021)Chen, Zhang, Kang, Zhe, Bao, Jia, and
  Lu]{chen2021animatable}
Chen, J., Zhang, Y., Kang, D., Zhe, X., Bao, L., Jia, X., and Lu, H.
\newblock Animatable neural radiance fields from monocular rgb videos.
\newblock \emph{arXiv preprint arXiv:2106.13629}, 2021.

\bibitem[Dou et~al.(2016)Dou, Khamis, Degtyarev, Davidson, Fanello, Kowdle,
  Escolano, Rhemann, Kim, Taylor, et~al.]{dou2016fusion4d}
Dou, M., Khamis, S., Degtyarev, Y., Davidson, P., Fanello, S.~R., Kowdle, A.,
  Escolano, S.~O., Rhemann, C., Kim, D., Taylor, J., et~al.
\newblock Fusion4d: Real-time performance capture of challenging scenes.
\newblock \emph{ACM TOG}, 35\penalty0 (4):\penalty0 1--13, 2016.

\bibitem[Dou et~al.(2017)Dou, Davidson, Fanello, Khamis, Kowdle, Rhemann,
  Tankovich, and Izadi]{dou2017motion2fusion}
Dou, M., Davidson, P., Fanello, S.~R., Khamis, S., Kowdle, A., Rhemann, C.,
  Tankovich, V., and Izadi, S.
\newblock Motion2fusion: Real-time volumetric performance capture.
\newblock \emph{ACM TOG}, 36\penalty0 (6):\penalty0 1--16, 2017.

\bibitem[Edelsbrunner et~al.(1983)Edelsbrunner, Kirkpatrick, and
  Seidel]{edelsbrunner1983shape}
Edelsbrunner, H., Kirkpatrick, D., and Seidel, R.
\newblock On the shape of a set of points in the plane.
\newblock \emph{TIT}, 29\penalty0 (4):\penalty0 551--559, 1983.

\bibitem[Geng et~al.(2023)Geng, Peng, Xu, Bao, and Zhou]{instant_nvr}
Geng, C., Peng, S., Xu, Z., Bao, H., and Zhou, X.
\newblock Learning neural volumetric representations of dynamic humans in
  minutes.
\newblock In \emph{CVPR}, 2023.

\bibitem[Guo et~al.(2017)Guo, Xu, Yu, Liu, Dai, and Liu]{guo2017real}
Guo, K., Xu, F., Yu, T., Liu, X., Dai, Q., and Liu, Y.
\newblock Real-time geometry, albedo, and motion reconstruction using a single
  rgb-d camera.
\newblock \emph{ACM TOG}, 36\penalty0 (4):\penalty0 1, 2017.

\bibitem[Guo et~al.(2019)Guo, Lincoln, Davidson, Busch, Yu, Whalen, Harvey,
  Orts-Escolano, Pandey, Dourgarian, et~al.]{guo2019relightables}
Guo, K., Lincoln, P., Davidson, P., Busch, J., Yu, X., Whalen, M., Harvey, G.,
  Orts-Escolano, S., Pandey, R., Dourgarian, J., et~al.
\newblock The relightables: Volumetric performance capture of humans with
  realistic relighting.
\newblock \emph{ACM TOG}, 38\penalty0 (6):\penalty0 1--19, 2019.

\bibitem[He et~al.(2020)He, Collomosse, Jin, and Soatto]{he2020geo}
He, T., Collomosse, J., Jin, H., and Soatto, S.
\newblock Geo-pifu: Geometry and pixel aligned implicit functions for
  single-view human reconstruction.
\newblock \emph{NeurIPS}, 33:\penalty0 9276--9287, 2020.

\bibitem[He et~al.(2021)He, Xu, Saito, Soatto, and Tung]{he2021arch++}
He, T., Xu, Y., Saito, S., Soatto, S., and Tung, T.
\newblock Arch++: Animation-ready clothed human reconstruction revisited.
\newblock In \emph{ICCV}, pp.\  11046--11056, 2021.

\bibitem[Hu \& Liu(2023)Hu and Liu]{hu2023gauhuman}
Hu, S. and Liu, Z.
\newblock Gauhuman: Articulated gaussian splatting from monocular human videos.
\newblock \emph{arXiv preprint arXiv:2312.02973}, 2023.

\bibitem[Huang et~al.(2020)Huang, Xu, Lassner, Li, and Tung]{huang2020arch}
Huang, Z., Xu, Y., Lassner, C., Li, H., and Tung, T.
\newblock Arch: Animatable reconstruction of clothed humans.
\newblock In \emph{CVPR}, pp.\  3093--3102, 2020.

\bibitem[Izadi et~al.(2011)Izadi, Kim, Hilliges, Molyneaux, Newcombe, Kohli,
  Shotton, Hodges, Freeman, Davison, et~al.]{izadi2011kinectfusion}
Izadi, S., Kim, D., Hilliges, O., Molyneaux, D., Newcombe, R., Kohli, P.,
  Shotton, J., Hodges, S., Freeman, D., Davison, A., et~al.
\newblock Kinectfusion: real-time 3d reconstruction and interaction using a
  moving depth camera.
\newblock In \emph{UIST}, pp.\  559--568, 2011.

\bibitem[Jiang et~al.(2022{\natexlab{a}})Jiang, Hong, Bao, and
  Zhang]{jiang2022selfrecon}
Jiang, B., Hong, Y., Bao, H., and Zhang, J.
\newblock Selfrecon: Self reconstruction your digital avatar from monocular
  video.
\newblock In \emph{CVPR}, pp.\  5605--5615, 2022{\natexlab{a}}.

\bibitem[Jiang et~al.(2023)Jiang, Chen, Song, and
  Hilliges]{jiang2023instantavatar}
Jiang, T., Chen, X., Song, J., and Hilliges, O.
\newblock Instantavatar: Learning avatars from monocular video in 60 seconds.
\newblock In \emph{CVPR}, pp.\  16922--16932, 2023.

\bibitem[Jiang et~al.(2022{\natexlab{b}})Jiang, Yi, Samei, Tuzel, and
  Ranjan]{jiang2022neuman}
Jiang, W., Yi, K.~M., Samei, G., Tuzel, O., and Ranjan, A.
\newblock Neuman: Neural human radiance field from a single video.
\newblock In \emph{ECCV}, pp.\  402--418. Springer, 2022{\natexlab{b}}.

\bibitem[Jung et~al.(2023)Jung, Brasch, Song, Perez-Pellitero, Zhou, Li, Navab,
  and Busam]{jung2023deformable}
Jung, H., Brasch, N., Song, J., Perez-Pellitero, E., Zhou, Y., Li, Z., Navab,
  N., and Busam, B.
\newblock Deformable 3d gaussian splatting for animatable human avatars.
\newblock \emph{arXiv preprint arXiv:2312.15059}, 2023.

\bibitem[Kanazawa et~al.(2018)Kanazawa, Black, Jacobs, and
  Malik]{kanazawa2018end}
Kanazawa, A., Black, M.~J., Jacobs, D.~W., and Malik, J.
\newblock End-to-end recovery of human shape and pose.
\newblock In \emph{CVPR}, pp.\  7122--7131, 2018.

\bibitem[Kerbl et~al.(2023)Kerbl, Kopanas, Leimk{\"u}hler, and
  Drettakis]{kerbl20233d}
Kerbl, B., Kopanas, G., Leimk{\"u}hler, T., and Drettakis, G.
\newblock 3d gaussian splatting for real-time radiance field rendering.
\newblock \emph{ACM TOG}, 42\penalty0 (4), 2023.

\bibitem[Kirillov et~al.(2023)Kirillov, Mintun, Ravi, Mao, Rolland, Gustafson,
  Xiao, Whitehead, Berg, Lo, et~al.]{kirillov2023segment}
Kirillov, A., Mintun, E., Ravi, N., Mao, H., Rolland, C., Gustafson, L., Xiao,
  T., Whitehead, S., Berg, A.~C., Lo, W.-Y., et~al.
\newblock Segment anything.
\newblock \emph{arXiv preprint arXiv:2304.02643}, 2023.

\bibitem[Kocabas et~al.(2020)Kocabas, Athanasiou, and Black]{kocabas2020vibe}
Kocabas, M., Athanasiou, N., and Black, M.~J.
\newblock Vibe: Video inference for human body pose and shape estimation.
\newblock In \emph{CVPR}, pp.\  5253--5263, 2020.

\bibitem[Kolotouros et~al.(2019)Kolotouros, Pavlakos, and
  Daniilidis]{kolotouros2019convolutional}
Kolotouros, N., Pavlakos, G., and Daniilidis, K.
\newblock Convolutional mesh regression for single-image human shape
  reconstruction.
\newblock In \emph{CVPR}, pp.\  4501--4510, 2019.

\bibitem[Kwon et~al.(2021)Kwon, Kim, Ceylan, and Fuchs]{kwon2021neural}
Kwon, Y., Kim, D., Ceylan, D., and Fuchs, H.
\newblock Neural human performer: Learning generalizable radiance fields for
  human performance rendering.
\newblock \emph{NeurIPS}, 34:\penalty0 24741--24752, 2021.

\bibitem[Lei et~al.(2024)Lei, Wang, Pavlakos, Liu, and Daniilidis]{lei2023gart}
Lei, J., Wang, Y., Pavlakos, G., Liu, L., and Daniilidis, K.
\newblock Gart: Gaussian articulated template models.
\newblock In \emph{CVPR}, 2024.

\bibitem[Li et~al.(2023{\natexlab{a}})Li, Bian, Xu, Chen, Yang, and
  Lu]{li2023hybrik}
Li, J., Bian, S., Xu, C., Chen, Z., Yang, L., and Lu, C.
\newblock Hybrik-x: Hybrid analytical-neural inverse kinematics for whole-body
  mesh recovery.
\newblock \emph{arXiv preprint arXiv:2304.05690}, 2023{\natexlab{a}}.

\bibitem[Li et~al.(2023{\natexlab{b}})Li, Tao, Yang, and Yang]{li2023human101}
Li, M., Tao, J., Yang, Z., and Yang, Y.
\newblock Human101: Training 100+ fps human gaussians in 100s from 1 view.
\newblock \emph{arXiv preprint arXiv:2312.15258}, 2023{\natexlab{b}}.

\bibitem[Lin et~al.(2023)Lin, Zeng, Wang, Zhang, and Li]{lin2023one}
Lin, J., Zeng, A., Wang, H., Zhang, L., and Li, Y.
\newblock One-stage 3d whole-body mesh recovery with component aware
  transformer.
\newblock In \emph{CVPR}, pp.\  21159--21168, 2023.

\bibitem[Lin et~al.(2021)Lin, Wang, and Liu]{lin2021end}
Lin, K., Wang, L., and Liu, Z.
\newblock End-to-end human pose and mesh reconstruction with transformers.
\newblock In \emph{CVPR}, pp.\  1954--1963, 2021.

\bibitem[Loper et~al.(2023)Loper, Mahmood, Romero, Pons-Moll, and
  Black]{loper2023smpl}
Loper, M., Mahmood, N., Romero, J., Pons-Moll, G., and Black, M.~J.
\newblock Smpl: A skinned multi-person linear model.
\newblock In \emph{Seminal Graphics Papers: Pushing the Boundaries, Volume 2},
  pp.\  851--866. 2023.

\bibitem[Ma et~al.(2020)Ma, Yang, Ranjan, Pujades, Pons-Moll, Tang, and
  Black]{ma2020learning}
Ma, Q., Yang, J., Ranjan, A., Pujades, S., Pons-Moll, G., Tang, S., and Black,
  M.~J.
\newblock Learning to dress 3d people in generative clothing.
\newblock In \emph{CVPR}, pp.\  6469--6478, 2020.

\bibitem[Ma et~al.(2021)Ma, Yang, Tang, and Black]{ma2021power}
Ma, Q., Yang, J., Tang, S., and Black, M.~J.
\newblock The power of points for modeling humans in clothing.
\newblock In \emph{ICCV}, pp.\  10974--10984, 2021.

\bibitem[Mildenhall et~al.(2021)Mildenhall, Srinivasan, Tancik, Barron,
  Ramamoorthi, and Ng]{mildenhall2021nerf}
Mildenhall, B., Srinivasan, P.~P., Tancik, M., Barron, J.~T., Ramamoorthi, R.,
  and Ng, R.
\newblock Nerf: Representing scenes as neural radiance fields for view
  synthesis.
\newblock \emph{Communications of the ACM}, 65\penalty0 (1):\penalty0 99--106,
  2021.

\bibitem[Newcombe et~al.(2015)Newcombe, Fox, and
  Seitz]{newcombe2015dynamicfusion}
Newcombe, R.~A., Fox, D., and Seitz, S.~M.
\newblock Dynamicfusion: Reconstruction and tracking of non-rigid scenes in
  real-time.
\newblock In \emph{CVPR}, pp.\  343--352, 2015.

\bibitem[Pavlakos et~al.(2019)Pavlakos, Choutas, Ghorbani, Bolkart, Osman,
  Tzionas, and Black]{SMPL-X:2019}
Pavlakos, G., Choutas, V., Ghorbani, N., Bolkart, T., Osman, A. A.~A., Tzionas,
  D., and Black, M.~J.
\newblock Expressive body capture: {3D} hands, face, and body from a single
  image.
\newblock In \emph{CVPR}, pp.\  10975--10985, 2019.

\bibitem[Peng et~al.(2021{\natexlab{a}})Peng, Dong, Wang, Zhang, Shuai, Zhou,
  and Bao]{peng2021animatable}
Peng, S., Dong, J., Wang, Q., Zhang, S., Shuai, Q., Zhou, X., and Bao, H.
\newblock Animatable neural radiance fields for modeling dynamic human bodies.
\newblock In \emph{ICCV}, pp.\  14314--14323, 2021{\natexlab{a}}.

\bibitem[Peng et~al.(2021{\natexlab{b}})Peng, Zhang, Xu, Wang, Shuai, Bao, and
  Zhou]{peng2021neural}
Peng, S., Zhang, Y., Xu, Y., Wang, Q., Shuai, Q., Bao, H., and Zhou, X.
\newblock Neural body: Implicit neural representations with structured latent
  codes for novel view synthesis of dynamic humans.
\newblock In \emph{CVPR}, pp.\  9054--9063, 2021{\natexlab{b}}.

\bibitem[Peng et~al.(2024)Peng, Xu, Dong, Wang, Zhang, Shuai, Bao, and
  Zhou]{peng2024animatable}
Peng, S., Xu, Z., Dong, J., Wang, Q., Zhang, S., Shuai, Q., Bao, H., and Zhou,
  X.
\newblock Animatable implicit neural representations for creating realistic
  avatars from videos.
\newblock \emph{IEEE TPAMI}, 2024.

\bibitem[Qian et~al.(2023)Qian, Kirschstein, Schoneveld, Davoli, Giebenhain,
  and Nie{\ss}ner]{qian2023gaussianavatars}
Qian, S., Kirschstein, T., Schoneveld, L., Davoli, D., Giebenhain, S., and
  Nie{\ss}ner, M.
\newblock Gaussianavatars: Photorealistic head avatars with rigged 3d
  gaussians.
\newblock \emph{arXiv preprint arXiv:2312.02069}, 2023.

\bibitem[Qian et~al.(2024)Qian, Wang, Mihajlovic, Geiger, and
  Tang]{qian20233dgs}
Qian, Z., Wang, S., Mihajlovic, M., Geiger, A., and Tang, S.
\newblock 3dgs-avatar: Animatable avatars via deformable 3d gaussian splatting.
\newblock 2024.

\bibitem[Remelli et~al.(2022)Remelli, Bagautdinov, Saito, Wu, Simon, Wei, Guo,
  Cao, Prada, Saragih, et~al.]{remelli2022drivable}
Remelli, E., Bagautdinov, T., Saito, S., Wu, C., Simon, T., Wei, S.-E., Guo,
  K., Cao, Z., Prada, F., Saragih, J., et~al.
\newblock Drivable volumetric avatars using texel-aligned features.
\newblock In \emph{ACM SIGGRAPH}, pp.\  1--9, 2022.

\bibitem[Saito et~al.(2019)Saito, Huang, Natsume, Morishima, Kanazawa, and
  Li]{saito2019pifu}
Saito, S., Huang, Z., Natsume, R., Morishima, S., Kanazawa, A., and Li, H.
\newblock Pifu: Pixel-aligned implicit function for high-resolution clothed
  human digitization.
\newblock In \emph{ICCV}, pp.\  2304--2314, 2019.

\bibitem[Saito et~al.(2020)Saito, Simon, Saragih, and Joo]{saito2020pifuhd}
Saito, S., Simon, T., Saragih, J., and Joo, H.
\newblock Pifuhd: Multi-level pixel-aligned implicit function for
  high-resolution 3d human digitization.
\newblock In \emph{CVPR}, pp.\  84--93, 2020.

\bibitem[Saito et~al.(2024)Saito, Schwartz, Simon, Li, and
  Nam]{saito2023relightable}
Saito, S., Schwartz, G., Simon, T., Li, J., and Nam, G.
\newblock Relightable gaussian codec avatars.
\newblock In \emph{CVPR}, 2024.

\bibitem[Simonyan \& Zisserman(2015)Simonyan and Zisserman]{SimonyanZ14a}
Simonyan, K. and Zisserman, A.
\newblock Very deep convolutional networks for large-scale image recognition.
\newblock In \emph{ICLR}, 2015.

\bibitem[Varol et~al.(2018)Varol, Ceylan, Russell, Yang, Yumer, Laptev, and
  Schmid]{varol2018bodynet}
Varol, G., Ceylan, D., Russell, B., Yang, J., Yumer, E., Laptev, I., and
  Schmid, C.
\newblock Bodynet: Volumetric inference of 3d human body shapes.
\newblock In \emph{ECCV}, pp.\  20--36, 2018.

\bibitem[Weng et~al.(2022)Weng, Curless, Srinivasan, Barron, and
  Kemelmacher-Shlizerman]{weng2022humannerf}
Weng, C.-Y., Curless, B., Srinivasan, P.~P., Barron, J.~T., and
  Kemelmacher-Shlizerman, I.
\newblock Humannerf: Free-viewpoint rendering of moving people from monocular
  video.
\newblock In \emph{CVPR}, pp.\  16210--16220, 2022.

\bibitem[Xiang et~al.(2020)Xiang, Prada, Wu, and
  Hodgins]{xiang2020monoclothcap}
Xiang, D., Prada, F., Wu, C., and Hodgins, J.
\newblock Monoclothcap: Towards temporally coherent clothing capture from
  monocular rgb video.
\newblock In \emph{3DV}, pp.\  322--332. IEEE, 2020.

\bibitem[Xie et~al.(2024)Xie, Zong, Qiu, Li, Feng, Yang, and
  Jiang]{xie2023physgaussian}
Xie, T., Zong, Z., Qiu, Y., Li, X., Feng, Y., Yang, Y., and Jiang, C.
\newblock Physgaussian: Physics-integrated 3d gaussians for generative
  dynamics.
\newblock In \emph{CVPR}, 2024.

\bibitem[Xiu et~al.(2022)Xiu, Yang, Tzionas, and Black]{xiu2022icon}
Xiu, Y., Yang, J., Tzionas, D., and Black, M.~J.
\newblock Icon: Implicit clothed humans obtained from normals.
\newblock In \emph{CVPR}, pp.\  13286--13296. IEEE, 2022.

\bibitem[Xiu et~al.(2023)Xiu, Yang, Cao, Tzionas, and Black]{xiu2022econ}
Xiu, Y., Yang, J., Cao, X., Tzionas, D., and Black, M.~J.
\newblock {ECON: Explicit Clothed humans Optimized via Normal integration}.
\newblock In \emph{CVPR}, 2023.

\bibitem[Yu et~al.(2021)Yu, Ye, Tancik, and Kanazawa]{yu2021pixelNeRF}
Yu, A., Ye, V., Tancik, M., and Kanazawa, A.
\newblock {pixelNeRF}: Neural radiance fields from one or few images.
\newblock In \emph{CVPR}, 2021.

\bibitem[Yu et~al.(2017)Yu, Guo, Xu, Dong, Su, Zhao, Li, Dai, and
  Liu]{yu2017bodyfusion}
Yu, T., Guo, K., Xu, F., Dong, Y., Su, Z., Zhao, J., Li, J., Dai, Q., and Liu,
  Y.
\newblock Bodyfusion: Real-time capture of human motion and surface geometry
  using a single depth camera.
\newblock In \emph{ICCV}, pp.\  910--919, 2017.

\bibitem[Yu et~al.(2018)Yu, Zheng, Guo, Zhao, Dai, Li, Pons-Moll, and
  Liu]{yu2018doublefusion}
Yu, T., Zheng, Z., Guo, K., Zhao, J., Dai, Q., Li, H., Pons-Moll, G., and Liu,
  Y.
\newblock Doublefusion: Real-time capture of human performances with inner body
  shapes from a single depth sensor.
\newblock In \emph{CVPR}, pp.\  7287--7296, 2018.

\bibitem[Yuan et~al.(2023)Yuan, Li, Huang, De~Mello, Nagano, Kautz, and
  Iqbal]{yuan2023gavatar}
Yuan, Y., Li, X., Huang, Y., De~Mello, S., Nagano, K., Kautz, J., and Iqbal, U.
\newblock Gavatar: Animatable 3d gaussian avatars with implicit mesh learning.
\newblock \emph{arXiv preprint arXiv:2312.11461}, 2023.

\bibitem[Zhang et~al.(2023)Zhang, Tian, Zhang, Li, An, Sun, and
  Liu]{zhang2023pymaf}
Zhang, H., Tian, Y., Zhang, Y., Li, M., An, L., Sun, Z., and Liu, Y.
\newblock Pymaf-x: Towards well-aligned full-body model regression from
  monocular images.
\newblock \emph{IEEE TPAMI}, 2023.

\bibitem[Zhang et~al.(2018)Zhang, Isola, Efros, Shechtman, and
  Wang]{zhang2018perceptual}
Zhang, R., Isola, P., Efros, A.~A., Shechtman, E., and Wang, O.
\newblock The unreasonable effectiveness of deep features as a perceptual
  metric.
\newblock In \emph{CVPR}, 2018.

\bibitem[Zheng et~al.(2021)Zheng, Yu, Liu, and Dai]{zheng2021pamir}
Zheng, Z., Yu, T., Liu, Y., and Dai, Q.
\newblock Pamir: Parametric model-conditioned implicit representation for
  image-based human reconstruction.
\newblock \emph{IEEE TPAMI}, 44\penalty0 (6):\penalty0 3170--3184, 2021.

\bibitem[Zhou et~al.(2021)Zhou, Habermann, Habibie, Tewari, Theobalt, and
  Xu]{zhou2021monocular}
Zhou, Y., Habermann, M., Habibie, I., Tewari, A., Theobalt, C., and Xu, F.
\newblock Monocular real-time full body capture with inter-part correlations.
\newblock In \emph{CVPR}, pp.\  4811--4822, 2021.

\bibitem[Zielonka et~al.(2023)Zielonka, Bagautdinov, Saito, Zollh{\"o}fer,
  Thies, and Romero]{zielonka2023drivable}
Zielonka, W., Bagautdinov, T., Saito, S., Zollh{\"o}fer, M., Thies, J., and
  Romero, J.
\newblock Drivable 3d gaussian avatars.
\newblock \emph{arXiv preprint arXiv:2311.08581}, 2023.

\end{thebibliography}
